\documentclass[3p,times]{elsarticle}

\usepackage{ecrc}

\usepackage{amssymb}

\usepackage{graphicx}
\usepackage{float} 
\usepackage{subfigure}
\usepackage{algorithm}
\usepackage{algpseudocode}
\usepackage{color}
\usepackage{makecell}
\usepackage{array}
\usepackage{booktabs} 
\usepackage{bbding}
\usepackage{pifont}
\usepackage{url}

\usepackage{marvosym}
\usepackage{amsmath}

\usepackage[table,xcdraw]{xcolor}

\volume{00}

\firstpage{1}

\journalname{Procedia Computer Science}

\runauth{}

\jid{procs}

\jnltitlelogo{Procedia Computer Science}

\CopyrightLine{2011}{Published by Elsevier Ltd.}

\usepackage[figuresright]{rotating}

\begin{document}

\begin{frontmatter}



\dochead{}

\title{FusionPlanner: A Multi-task Motion Planner for Mining Trucks \\ via Multi-sensor Fusion}


\author[1,2]{Siyu Teng*}
\author[1,2]{Luxi Li*}
\author[1,2]{Yuchen Li}
\author[3]{Xuemin Hu}
\author[4,5]{Lingxi Li}
\author[4,6]{\\Yunfeng Ai\textsuperscript{\Letter}}
\author[4,7]{Long Chen\textsuperscript{\Letter}}
\address[1]{Department of Computer Science, Hong Kong Baptist University, Hong Kong SAR, China}
\address[2]{Department of Computer Science, BNU-HKBU United International College, Zhuhai, China}
\address[3]{The School of Artificial Intelligence, Hubei University, Wuhan, China}
\address[4]{Waytous Inc., Haidian Distinct, Beijing, China}
\address[5]{Department of Electrical and Computer Engineering, Indiana University-Purdue University Indianapolis (IUPUI), Indianapolis, USA}
\address[6]{School of Artificial Intelligence, University of Chinese Academy of Sciences, Beijing, China}
\address[7]{State Key Laboratory of Management and Control for Complex Systems, Institute of Automation, Chinese Academy of Sciences, China}


\footnote{* Equal contribution}

\footnote{Yunfeng Ai\textsuperscript{\Letter} and Long Chen\textsuperscript{\Letter} (long.chen@ic.ac.cn) are corresponding authors.}

\footnote{This research is the result of the research project funded by the National Natural Science Foundation of China under Grant 62373356 and 62273135; the Natural Science Foundation of Hubei Province in China under Grant 2021CFB460}

\begin{abstract}

In recent years, significant achievements have been made in motion planning for intelligent vehicles. However, as a typical unstructured environment, open-pit mining attracts limited attention due to its complex operational conditions and adverse environmental factors. A comprehensive paradigm for unmanned transportation in open-pit mines is proposed in this research. \textcolor{black}{Firstly,  we propose a multi-task motion planning algorithm, called FusionPlanner, for autonomous mining trucks by the multi-sensor fusion method to adapt both lateral and longitudinal control tasks for unmanned transportation. Then, we develop a novel benchmark called MiningNav, which offers three validation approaches to evaluate the trustworthiness and robustness of well-trained algorithms in transportation roads of open-pit mines. Finally, we introduce the Parallel Mining Simulator (PMS), a new high-fidelity simulator specifically designed for open-pit mining scenarios. PMS enables the users to manage and control open-pit mine transportation from both the single-truck control and multi-truck scheduling perspectives.} \textcolor{black}{The performance of FusionPlanner is tested by MiningNav in PMS, and the empirical results demonstrate a significant reduction in the number of collisions and takeovers of our planner. We anticipate our unmanned transportation paradigm will bring mining trucks one step closer to trustworthiness and robustness in continuous round-the-clock unmanned transportation.}
\end{abstract}


\end{frontmatter}


\section{Introduction}
\label{Intro}


With the continual advancements in computational capabilities and the iterative progress of learning methodologies, automation has made a profound impact on various sectors of society \cite{Xia_1}. The mining industry encounters a multitude of challenges, including fluctuations in mineral prices, limited automation progress, and adverse working conditions. Moreover, the industry struggles with pressing issues such as an aging workforce, a shortage of technicians, rising labor costs, and increasingly stringent environmental regulations. Collectively, these factors present formidable obstacles to the development of automation in the mining industry \cite{zhang2023}.

Open-pit mining serves as the primary method for mineral extraction, focusing on minerals located near the surface and avoiding the costly infrastructure associated with underground mining. Open-pit mining assumes a crucial role in meeting global mineral demands and supplying essential raw materials across diverse industries. The expansive surface space facilitates the utilization of abundant large-scale specialized machinery, such as mining trucks, electric shovels, wheel loaders, and bulldozers, thereby significantly augmenting production efficiency. Furthermore, the availability of a multitude of controllable specialized machinery establishes a strong foundation for unmanned mining operations.


\begin{figure}[H]
    \centering
    \includegraphics[width=0.9\linewidth]{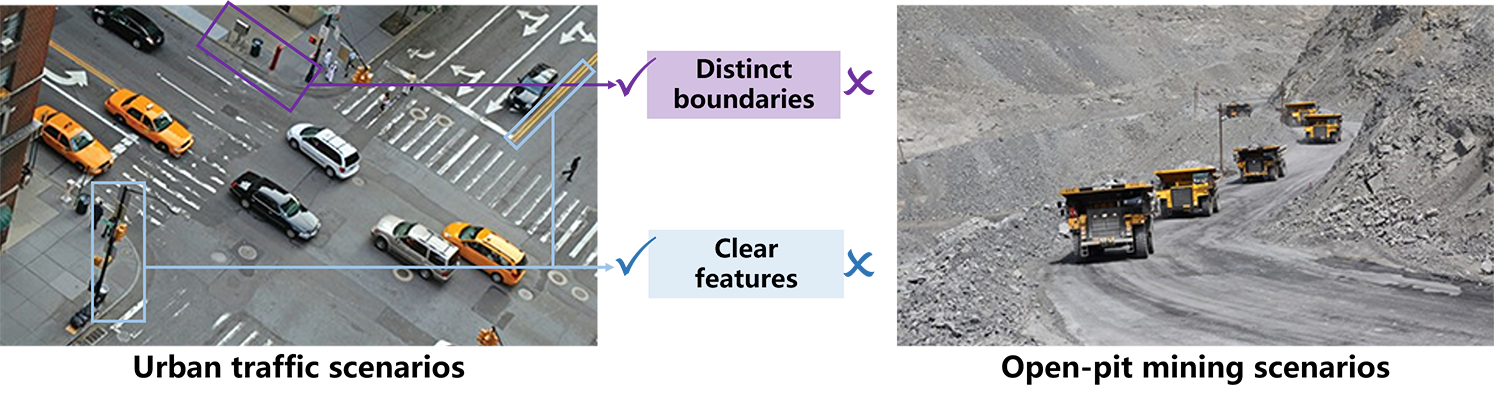}
    \caption{ The difference between open-pit mining scenarios and urban traffic scenarios.}
    \label{fig:compare}
\end{figure}



\textcolor{black}{Open-pit mines stand for a prototypical unstructured scenario, exhibiting numerous latent challenges unencountered within urban environments, as illustrated in Fig. \ref{fig:compare}. Urban environments present distinct boundary information, while the retaining walls and drivable areas in open-pit mines are always composed of homogenous materials, without distinct boundaries. Urban environments incorporate plenty of traffic elements and scenario information, affording a profusion of features, conversely, within open-pit mines, transportation participants manifest fewer categories, coupled with the absence of heterogeneous environmental contexts, engendering scarcity of clear features. }Mining trucks, the most efficient transportation specialized equipment in open-pit mines, provide a reasonable and feasible solution to the above problems with their intelligence: \textcolor{black}{1). Autonomous mining trucks contribute to transportation safety. In operation environments, such as loading and dumping sites, involving coordinated operations of large-scale specialty machinery, exist inherent risks for workers. These trucks effectively minimize human operation errors to mitigate risk status with other transportation participants. 2). Autonomous mining trucks greatly improve transportation efficiency, which demonstrates exceptional execution efficiency and precise control capabilities, allowing them to efficiently carry out various transportation tasks. Equipped with non-visual sensors such as LiDAR and millimeter-wave radar, these trucks can operate continuously around the clock, further enhancing the transportation efficiency of open-pit mines.}

\textcolor{black}{Autonomous mining trucks represent a significant advancement in open-pit mines, with wide-ranging impacts in terms of safety, productivity, cost control, and sustainable development \cite{Part0, Part1, Part2}. However, due to complex operational conditions and adverse environmental factors, the deployment of autonomous driving encounters considerable obstacles. Furthermore, the unstructured propriety of open-pit mines, characterized by limited effective features and blurred boundary information, poses unique challenges compared to structured urban scenarios. This research aims to address these challenges with an effective paradigm for unmanned transportation in open-pit mines, including an end-to-end motion planner, a novel benchmark, and a simulation environment.}

Currently, numerous end-to-end motion planners are proposed, however, to the best of our knowledge, there is no appropriate planner specifically designed for mining trucks. \textcolor{black}{On one hand, the advancement of end-to-end frameworks heavily relies on multi-sensor fusion. Some researchers \cite{Bojarski, CIL} combine images from three distinct perspectives of front-view cameras and tachometer data to predict vehicle control commands. Zhang et al. \cite{BEV,bevfusion} integrate target-level information from perception networks to further reduce the learning burden of the end-to-end planner to improve control accuracy. However, deploying the aforementioned algorithms in open-pit mines often proves impractical due to the formidable challenge of obtaining relevant features in real scenarios. Simultaneously, visual sensors \cite{bevfusion, CIL} relying on sufficient lighting may not meet the operational demands of open-pit mines, which require round-the-clock production. On the other hand, due to the unique conditions of open-pit mines, existing benchmarks \cite{carla, Zhu} cannot be directly applied to unmanned transportation, leading to a lack of a suitable benchmark for evaluating the trustworthiness and robustness of motion planners in open-pit mines. Furthermore, simulators serve as the primary platform for data collection and algorithm validation in end-to-end motion planners.} \textcolor{black}{However, due to the sparse textures and blurred edges of open-pit mines, feature extraction is a formidable task. Therefore, mainstream simulators designed for structured urban environments are not suitable for this situation.} 

In this research, we focus on investigating various issues in open-pit mines and establishing the first paradigm for unmanned transportation in mining trucks. Our contributions can be summarized as follows:


\begin{itemize}

\item [1)] \textcolor{black}{A multi-task motion planning algorithm is proposed, named FusionPlanner, which is specifically designed for mining trucks. It embeds the multi-task learning task to enhance the robustness of control and incorporates the hybrid evidential fusion model to enhance the trustworthiness of control commands.}

\item [2)] \textcolor{black}{A novel validation benchmark is constructed, designated as MiningNav, which is the first benchmark devised for unmanned transportation in open-pit mines. It presents three testing methods to assess the trustworthiness and robustness of algorithms, covering all transportation roads within open-pit mines.}

\item [3)] \textcolor{black}{A dedicated simulation platform is provided, called Parallel Mining Simulator (PMS), for unmanned transportation in open-pit mines, containing multiple sensors and transportation participants models. The platform also supports the configuration of mine scenarios, weather conditions, and other parametric variables.}

\end{itemize}

\section{Preliminaries}

\subsection{End-to-end Methods}

In general, autonomous driving frameworks are primarily divided into two categories: the relatively traditional pipeline framework, and the structurally innovative end-to-end framework.

\textcolor{black}{The pipeline framework is comprised of explicit submodels such as perception \cite{10227873, 10194984}, decision-making \cite{decision}, planning \cite{hu, planning1}, and control \cite{control1}, and each individual submodel can be optimized independently \cite{zhangbest1}.} \textcolor{black}{Owing to its clear intermediate representation and deterministic decision logic, the pipeline framework possesses the ability to infer how the framework computes specific control commands, when unpredicted behavior occurs, the pipeline can precisely locate the root of errors.} Therefore, numerous exceptional works are already been deployed in intelligent vehicles \cite{Li1}. Nevertheless, the pipeline framework has two fatal issues. Robustness issue: the failure of any individual submodel can result in a complete crash of the entire framework, and the most optimal submodel does not stand for optimality at the framework level \cite{XIA2016557}; Real-time issue: the concatenated architecture, along with excessive hand-craft heuristic functions have impeded the decision speed of the framework \cite{teng1}.

Confront the above issues in the pipeline framework, an arising end-to-end framework presents a novel solution \cite{teng2}. The end-to-end framework integrates the total driving task as a singular learning task, encompassing raw perception data to control commands \cite{10164220}. By predicting the optimal intermediate representation for the target task, this framework strengthens its generalizability in various complex scenarios. \textcolor{black}{The amazing architecture of the end-to-end framework only comprises one or several simple networks, allowing the framework to extract necessary implicit features from raw data without human-defined bottlenecks. Compared to the pipeline framework, it demonstrates exceptional real-time performance and extrapolation capabilities.} Bojarski et al. \cite{Bojarski} are pioneers in this field, successfully employing neural networks to learn the steering policy. However, this model is only applicable to the lane-keeping task in simple scenarios and struggles to adapt to complex urban environments. Codevilla et al. \cite{CIL} proposed a classical Conditional Imitation Learning (CIL), a classical algorithm that utilizes High-Level Commands (HIL) to select corresponding output branches, thereby enhancing the robustness of the motion planner. Additionally, CILRS \cite{speedbranch}, a derivative of CIL, integrates the correlation between tachometer data and the input perception data to improve control stability. Zhu et al. \cite{Zhu} decouples the vehicle control commands into lateral and longitudinal and implements imitation learning (IL) at complex urban traffic intersections through a multi-task learning approach.

In general, IL with an offline training algorithm has two critical challenges: 1). Covariate shift \cite{CascadedNet}, making IL difficult to generalize in dense traffic, and causal confusion \cite{confusion, XIA2022107993}, as models struggle to identify genuine causal relationships in experts demonstrations. Numerous algorithms have been proposed to alleviate the aforementioned problems. 2). Privileged supervision, such as route navigation \cite{nav, CDQN} or BEV map \cite{BEV, bevfusion}, is utilized as input perception data. Object-level detection results of traffic participants can be integrated into the perception input, reducing the feature extraction burden in motion planners \cite{10148929}.

While privileged information is easily available in simulators, extracting relevant features from real-world observations proves challenging, making it impractical to deploy such algorithms in the real world. To overcome the covariate shift problem, some researchers \cite{nav,27_2} employ DAgger \cite{30_2} to elevate the offline training of IL to online optimization. Furthermore, reinforcement learning (RL) can provide a solution by facilitating more online exploration, where the offline-trained IL agent initializes the model for the RL agent \cite{BEV, 9770186}. However, online RL is primarily applicable in simulators, and testing well-trained in the real world is challenging due to expensive costs. Moreover, a carefully designed reward function may not reflect human driving decisions. To overcome the difficulties in finding the optimal reward function, Inverse Reinforcement Learning (IRL) is proposed to learn human driving decisions \cite{39_2,XIA2021107290}. However, model-based IRL \cite{35_2} requires a perfect understanding of transition functions, while model-free IRL \cite{38_2} necessitates online interaction with the environment. Additionally, the computational burden of IRL may be heavy, as it often iteratively solves forward RL problems after calculating each new reward \cite{end2end, 10198672}.

Without exception, the aforementioned research is exclusively focused on structured urban scenarios, with very few addressing unstructured scenarios like open-pit mines \cite{10149821}. Firstly, the complex operational conditions and harsh environmental factors in open-pit mines make end-to-end motion planners designed for urban scenarios unsuitable for deployment. Secondly, mining trucks, as special machines in open-pit mines, exhibit significant differences in control logic compared to standard automobiles. Finally, the proposed motion planner in open-pit mines learns from demonstrations provided by human experts and autonomous driving agents, making a rigorously offline learning approach more suitable for real-world deployment. In this research, we present a trustworthiness and robustness imitation learning motion planning model, named FusionPlanner, specifically designed for mining trucks.

\subsection{Benchmarks}

\textcolor{black}{Benchmarks serve as a pivotal measure for evaluating the trustworthiness and robustness of end-to-end planners. The first native benchmark in urban scenarios is proposed within CARLA \cite{carla} and subsequently resolved with near-flawless scores. Subsequent benchmarks \cite{offline, prakash2021multi, coach, speedbranch} assure a stark contrast between training and testing scenarios, with the latter frequently demonstrating heightened challenge levels. The LAV benchmark \cite{chen2022learning} solely utilizes Town02 and Town05 for training, in a quest to maintain a diversity of test scenarios, thus leveraging all remaining scenarios for testing the well-trained planner. The Longest6 benchmark \cite{Longest} leverages six towns for end-to-end planner validation, while IntersectNav \cite{Zhu} shifts the focal point of testing to the intricate vehicular-pedestrian mixed intersections, thereby examining planner robustness amid densely populated crossroads. Two servers for the CARLA planner, namely the Leaderboards (v1 and v2), are published on the CARLA Autonomous Driving Leaderboard \cite{leaderboard} to corroborate the trustworthiness and robustness of the well-trained planner in genuine simulation scenarios. Leaderboards assure fair and comprehensive comparisons through random selection mechanisms. The Leaderboard v2 presents formidable challenges owing to its extensive route length (averaging over 8 kilometers), which inclusion of multiple novel traffic scenarios.}

\textcolor{black}{Despite the benchmarks aforementioned targeting city scenarios constructed by CARLA, the driving tasks and scenario features in unstructured roads such as open-pit mines exhibit significant discrepancies, rendering a straightforward standard transplant inapplicable. In this research, we provide the first validation benchmark of autonomous mining trucks, named MiningNav, for unmanned transportation in open-pit mines. }

\subsection{Driving Simulators}

\linespread{1.05}
\begin{table}[t]\footnotesize \centering 

\caption{Comparison of reviewed driving simulators in this research.}
\label{tab:simulator}
\begin{tabular}{m{1.9cm}<{\centering}cccccccc}
\toprule


Simulator & \multicolumn{1}{c}{\makecell{Vehicle\\Dynamics}} & \multicolumn{1}{c}{\makecell{Muti-agent\\Support}} & \multicolumn{1}{c}{\makecell{Custom\\Map}} & \multicolumn{1}{c}{\makecell{Lidar and\\ Camera}} & \makecell{Visual\\Rendering} & \makecell{Active\\Maintenance} & \makecell{Weather\\Conditions} & \makecell{Light\\Weight} \\ \hline

CARLA \cite{carla}     & \multicolumn{1}{c}{$\checkmark$}                                         & \multicolumn{1}{c}{$\checkmark$}                                           & \multicolumn{1}{c}{$\checkmark$}                                   & \multicolumn{1}{c}{$\checkmark$}                                     & $\checkmark$                          & $\checkmark$          & $\checkmark$         &             \\ \hline
SUMMIT \cite{4}   & \multicolumn{1}{c}{$\checkmark$}                                         & \multicolumn{1}{c}{$\checkmark$}                                           & \multicolumn{1}{c}{$\checkmark$}                                   & \multicolumn{1}{c}{$\checkmark$}                                     & $\checkmark$                          & $\checkmark$         & $\checkmark$          &             \\ \hline
MACAD \cite{34}    & \multicolumn{1}{c}{$\checkmark$}                                         & \multicolumn{1}{c}{$\checkmark$}                                           & \multicolumn{1}{c}{$\checkmark$}                                   & \multicolumn{1}{c}{$\checkmark$}                                     & $\checkmark$                          & $\checkmark$           &        &              \\ \hline
GTA V \cite{28}    & $\checkmark$                                                             &                                                                &                                                        & $\checkmark$                                                         & $\checkmark$                           &         & $\checkmark$           &             \\ \hline
SIM4CV \cite{31}    & $\checkmark$                                                             &                                                                &                                                        & $\checkmark$                                                         & $\checkmark$                           &          &          &             \\ \hline
TORCS  \cite{57}     & $\checkmark$                                                             &                                                                & $\checkmark$                                                       & $\checkmark$                                                         &                            &            &        & $\checkmark$            \\ \hline
DriverGym \cite{20} &                                                              &                                                                &                                                        &                                                          &                            &  $\checkmark$              &    & $\checkmark$            \\ \hline
FLOW \cite{55}     &                                                              &                                                                & $\checkmark$                                                       &                                                          &                            &           &         & $\checkmark$            \\ \hline
CityFlow \cite{60} &                                                              & $\checkmark$                                                               & $\checkmark$                                                       &                                                          &                            &        &            & $\checkmark$            \\ \hline
SUMO \cite{27}     &                                                              &                                                                & $\checkmark$                                                       &                                                          &                            & $\checkmark$             &      & $\checkmark$            \\ \hline
MADRaS \cite{42}    & $\checkmark$                                                             & $\checkmark$                                                               & $\checkmark$                                                       & $\checkmark$                                                         &                            &              &      & $\checkmark$            \\ \hline
PMS       & $\checkmark$                                                             & $\checkmark$                                                               & $\checkmark$                                                        & $\checkmark$                                                         & $\checkmark$               & $\checkmark$            & $\checkmark$                   &          \\ \bottomrule
\end{tabular}
\end{table}

In the past few years, due to considerations of public safety and economic costs, significant progress has been achieved in the field of autonomous driving simulation. Research indicates that intelligent vehicles must cover billions of miles to prove their trustworthiness and robustness, which would be impossible without the aid of simulation. Since the beginning of autonomous driving research, simulators have played a pivotal role in the development and validation process \cite{xia2022simulation}. They enable the validation of algorithms without the need for operating physical vehicles. 

Simulation offers several key advantages over on-road validation, particularly in dangerous scenarios \cite{XIA3}. It provides a safe and cost-effective digital environment, allowing for the exploration of rare corner cases that are infrequently encountered in the real world, such as extreme weather conditions. Furthermore, simulators accurately report all factors of a problematic scenario, facilitating effective debugging and validation of well-trained models. There exist numerous excellent autonomous driving simulators specialized for urban traffic scenarios, as shown in Tab. \ref{tab:simulator}. Simulators such as GTA V \cite{28}, Sim4CV \cite{31}, CARLA \cite{carla}, and their derivative projects SUMMIT \cite{4} and MACAD \cite{34} faithfully replicate the textures and boundaries of the physical world, including impressive effects in terms of lighting changing and weather effects. The high-fidelity rendering of these simulators establishes reliable and well-structured simulation environments.

\textcolor{black}{Despite the remarkable progress achieved in autonomous driving simulation, little attention is paid to open-pit mining scenarios. The unique characteristics and specialized machinery of open-pit mines make it impractical for structured scenario-based simulators to provide any form of algorithm validation. Consequently, the lack of a customized simulator remains a significant obstacle preventing the widespread deployment of autonomous driving in open-pit mines. To the best of our knowledge, there is no simulator specifically designed for the open-pit mine. Due to the complex operational conditions and harsh environmental factors, most simulators designed for structured scenarios fail to realistically represent unstructured environments. In this research, we develop a simulator named PMS for unmanned transportation in open-pit mines. PMS achieves a competitive level of environmental rendering capabilities for unstructured scenarios.}

For single-agent scenarios, most simulators, such as TORCS \cite{57} and DriverGym \cite{20}, only simulate vehicles with simple kinematic models, for example, bicycle models \cite{37}. This simplification allows for high-level discrete control tasks or provides simplified driving scenarios. To ensure more accurate control feedback in the simulated world of PMS, complex kinematics models are incorporated for all special machines, including mining trucks,  bulldozers, and excavators. The chassis, wheels, and joints of these machines are constructed under appropriate constraint conditions, ensuring that they closely resemble their real-world counterparts.

Regarding multi-agent scenarios, CityFlow \cite{60} and FLOW \cite{55} are two macroscopic traffic simulators based on SUMO \cite{27}, which primarily focus on simulating traffic flow but lack access ports to individual agent behavior information. MACAD \cite{34} and MADRaS \cite{42} are two traditional transportation simulators built on CARLA \cite{carla}, respectively. The newly developed SMARTS \cite{61} provides an excellent testing platform for agents and traffic participants in atomic traffic scenarios. In contrast, PMS offers a comprehensive perspective by encompassing the overall transportation scenario of the open-pit mine, while providing real-time feedback interfaces for individual vehicle states. Moreover, PMS supports custom map creation, allowing for the creation of complex open-pit mining operation scenarios such as loading and dumping sites. It can accommodate up to 300 trucks running simultaneously. Compared to existing simulators, our proposed simulator possesses essential features of composability and scalability, offering new research opportunities for autonomous driving applications in open-pit mining.

\section{FusionPlanner}

\begin{figure}[H]
    \centering
    \includegraphics[width=0.9\linewidth]{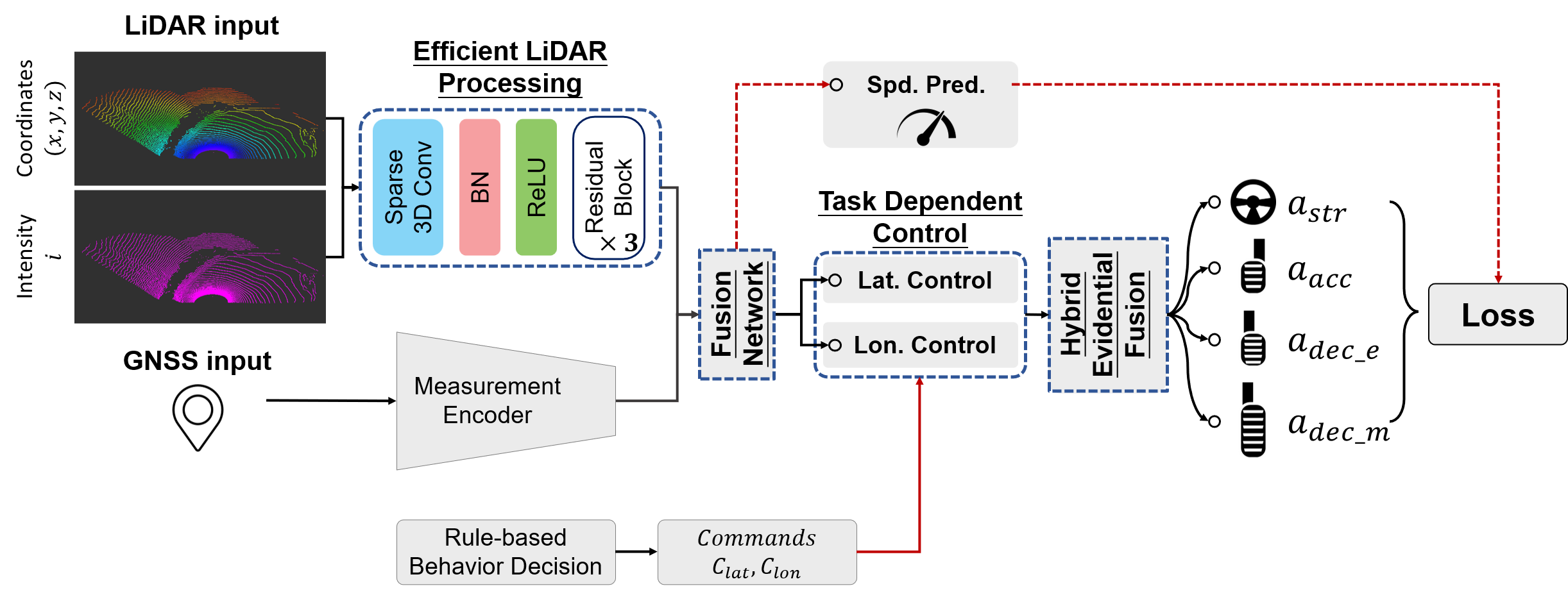}
    \caption{Overview of FusionPlanner. The raw LiDAR point cloud (with visual colorization based on coordination and intensity) and GNSS data are propagated through a fully connected layer in the control module of FusionPlanner after feature extraction.}
    \label{fig:framework}
\end{figure}


\textcolor{black}{Open-pit mines stand for a prototype illustration of the unstructured scenario, which is distinguished by its complex operational conditions and adverse environmental factors, thereby posing formidable obstacles to the progress of unmanned transportation in open-pit mines. This research focuses on the navigation task of the autonomous mining truck on transportation roads, where needs to interact safely with other transportation participants and predict the appropriate control commands to avoid impending accidents in a long driving distance (e.g. from the loading area to the unloading area). In order to complete the transportation task, the truck needs the execution of a series of driving commands. Specifically, lateral commands of mining trucks include three branches: go straight, turn left, and turn right. Longitudinal commands are also divided into three parts: accelerate, maintain, and decelerate.} Open-pit mines require round-the-clock production operations, due to its light-free environment at night, numerous visual sensors encounter operational impediments. \textcolor{black}{In this context, LiDAR serves as an indispensable sensor for autonomous mining trucks \cite{XiaLidar}. In addition, the integration of multiple data sources can significantly enhance robustness in the perception layer, which plays a crucial role in maintaining the accuracy of autonomous driving algorithms \cite{MSSP1,MSSP2,MSSP3}.} This research constructs the first end-to-end motion planner, named FusionPlanner, based on the fusion of LiDAR and GNSS for the mining truck, as shown in Fig.\ref{fig:framework}.

\textcolor{black}{CIL \cite{CIL} provides an outstanding framework for end-to-end autonomous driving, employing an instantaneous execution that spans from perception to control, which is also the main framework followed by this research. The proposition of CIL is formulated as follows: $\mathcal{D} = \left \{ \xi _{i}  \right \}^{N}_{i = 1}$, which stands for the $N$ trajectories collected from expert drivers and simulation programs. Each trajectory $\xi _{i}$ is  composed of sequences of observation-action pairs $\left \{ \left ( o_{i}^{t},a_{i}^{t},c_{i}^{t} \right )  \right \} _{t=1}^{T}$. The goal of this proposition is to learn the policy $\pi$ parameterized by $\theta$ that best fits the raw input observation to the output control command. The most optimal parameters $\theta^{*}$ are computed by minimizing the following loss function $\mathcal{L} $:}

\begin{eqnarray}
    \theta ^{*}=\mathop{\arg\min}\limits_{\theta} \sum_{j}\mathcal{L}\left ( \pi_{\theta} \left ( o_{j},c_{j}  \right ),a_{j}  \right )  
    \label{equ:lossfuncion}
\end{eqnarray}

\noindent
\textcolor{black}{$o_{i}^{t}$ stands for the raw observation data, which contains monocular front-view RGB Image $I_{i}^{t}$ and scalar value speed of ego-vehicle $v_{i}^{t}$; 
$a_{i}^{t}$ is a tuple that incorporates lateral commands $a_{i}^{t,str} \in [-1, 1]$ and longitudinal control commands $a_{i}^{t,acc} \in [-1, 1]$ of ego-vehicle; 
$c_{i}^{t}$ denotes the high-level command, which is to provide assistance to the ego-vehicle in selecting the most suitable branch, including go straight, follow lane, turn left, and turn right.}

The LiDAR output frequency ($\sim$10Hz) fails to meet the required control frequency ($>$50Hz) by mining trucks, to address this limitation, FusionPlanner is designed to predict multiple future control commands. Once the mining truck reaches these points, we can integrate and fuse these predictions, thereby stabilizing control. Furthermore, FusionPlanner takes a further step by fusing control predictions from different timestamps based on the uncertainty of the model at each point to deal with out-of-distribution (OOD) events or fuzzy future environments.

Given a raw LiDAR point cloud $o_{L}$ (\textit{coordinates} and \textit{intensity}) and raw localization data $o_{G}$ (\textit{longitude}, \textit{latitude}, and \textit{altitude}), the object of our motion planner is to learn an optimal policy $\pi_{\theta}$ with parameter $\theta$ by multi-sensor fusion method. The planner directly computes a sequence of control commands for the truck as well as the corresponding epistemic uncertainty for command filtering.

\begin{eqnarray}
    \left \{ \left ( a_{k},e_{k} \right )  \right \} _{k=0}^{K-1}=\pi_{\theta}(o_{L},o_{G},c),
    \label{equ: totlfunc}
\end{eqnarray}
where $K$ stands for the number of output predictions of the planner, each tuple of $a_{k}$ and $e_{k}$ represent a predicted control command with a lookahead distance of $k$[m] from the current frame: $a_{k}$ shows the computed control command, which can be supervised by the recorded label control command $y_{k}$, $a_{0}$ is the command for instantaneous execution, and $e_{k}$ are the hyperparameters to estimate the uncertainty associated with this prediction. $c$ shows high-level commands for lateral and longitudinal control.

The current control command $a_{0}$ can be used to control the truck and the remaining control commands $a_{k}, (k > 0)$ might be employed to enhance the robustness of the well-trained model through uncertainty-weighted temporal fusion using the hyperparameters $e_k$ ($k > 0$). By predicting additional control commands $a_{k}, (k > 0)$, FusionPlanner gains the ability to anticipate future scenarios and respond accordingly. The overall loss function of our model can be formulated as follow:

\begin{eqnarray}
    \theta ^{*}=\mathop{\arg\min}\limits_{\theta} \sum_{j}\mathcal{L}\left ( \pi_{\theta} \left ( o_{L},o_{G},c_{j}  \right ),\left \{ (y_{k}) \right \}_{k=0}^{K-1}  \right ),
    \label{equL: 8}
\end{eqnarray}
where the predicted control commands ${\left \{ (a_{k}) \right \} }_{k=0}^{K-1}$ are predicted from FusionPlanner with the policy $\pi_{\theta}$, and these commands are used to fit the labels commands ${\left \{ (y_{k}) \right \} }_{k=0}^{K-1}$ to calculate the overall loss $\mathcal{L}$. The main object of FusionPlanner is to calculate the most optimal parameters $\theta^{*}$ for the trustworthiness, robustness, and efficient end-to-end planning model in open-pit mines.

The following subsections outline the different components of FusionPlanner. Firstly, we demonstrate a novel method for extracting effective representations directly from the raw input point cloud data, along with GNSS data in Section \ref{sec:lidarprocess}. These representations serve as inputs to the fusion network, which predicts control commands and corresponding uncertainty estimates, as described in Section \ref{sec: fusionnetwork}. Subsequently, in Section \ref{sec:Task-dependentLos}, we introduce multi-task learning of lateral and longitudinal control for mining trucks, which offers a specialized parameter adaptive method, enabling FusionPlanner to autonomously adjust the weights of lateral and longitudinal control commands based on scenario variations in open-pit mines. Finally, we describe the optimization of our planner to extract features of inherent uncertainty in Section \ref{sec:evidentialfusion}, presenting a novel method for leveraging this uncertainty, ultimately increasing the trustworthiness of the autonomous truck.

\subsection{Efficient LiDAR Processing}\label{sec:lidarprocess}
Most recent 3D neural networks employ sparse convolution to handle point clouds, owing to their sparsity and disorderliness \cite{LiuBook}. Sparse convolution stores and computes only at non-zero positions, thereby conserving memory consumption and computational resources. Several optimization measures are proposed to accelerate sparse CUDA kernels. Firstly, the index lookup of sparse convolution, which queries the index of a given coordinate, is typically implemented by a hash map on a multi-threaded CPU \cite{MinCNN}. This bottleneck is limited by the parallelism of the CPU. To address this problem, Liu et al. \cite{Liu} maps the hash-based index lookup implementation to the GPU and utilizes Cuckoo hashing \cite{cuckoo} due to its good parallelism in both construction and querying. This optimization effectively reduces the overhead of kernel graph construction to just 10$\%$. Secondly, sparse computation is affected by the high cost of irregular memory access since neighboring points in 3D spaces may not be adjacent in the sparse tensor. To overcome this problem, Tang et al. \cite{Engine} aggregates memory access before convolution through an aggregation procedure and performs matrix multiplication on a large regular matrix.

Apart from optimizing sparse convolutions from the perspective of computational efficiency at the low level of computer operations, we further improve the computing speed of the feature extraction process from point clouds by redesigning the sparse convolution network from \cite{Liu}. For 2D CNNs, channel pruning is a commonly employed method to reduce the latency \cite{45, 46}. Nonetheless, as the sparse convolution is primarily constrained by data movement, rather than matrix multiplication, reducing the channel numbers is less effective. In this research, we adopt the approach proposed in \cite{Liu}, which proposes a comprehensive reduction strategy including channel numbers and network depths, rather than only focusing on pruning the channel dimension. The chosen dimensions exhibit a strong linear correlation with both the input points (resolution) and the number of layers (depth), resulting in significant improvements in both computation and memory access costs.

In light of the aforementioned, as shown in Fig.\ref{fig:framework}, our LiDAR feature extraction network operates on the input with the voxel size of 0.2m and comprises three residual sparse convolution blocks to reinforce features and reduce degradation, each block containing 16, 32, and 64 channels respectively. Additionally, two 3D sparse convolutions are employed to extract local features. To further enhance the perceptive field, we incorporate an additional sparse convolution following each block with the stride of 2.

\subsection{Fusion Network} \label{sec: fusionnetwork}
\textcolor{black}{The perception from a single sensor often lacks sufficient information in specific environments, resulting in low trustworthiness and robustness of well-trained models. multi-sensor fusion combines perception data from multiple sources across different levels and spaces with the unique characteristics of each sensor, creating a unified representation of the observed scenario. By integrating redundant data, multi-sensor fusion enhances the trustworthiness and robustness of feature representation, ensuring the safety of autonomous driving to the fullest extent.}

\textcolor{black}{Similar to previous works \cite{Bojarski, CIL,BEV,bevfusion}, FusionPlanner also inherits the advantages of multi-sensor fusion, employing a post-fusion approach to integrate extracted presentation from LiDAR and GNSS.} As shown in Fig.\ref{fig:framework}, the measurement Encoder in our planner consists of two fully connected networks, each with 512 units. This dual connection enhances the extraction of GNSS features, allowing for a more comprehensive feature representation of the input data. The latent features are compressed into 256 units in the final layer. The Fusion Network concatenates the latent features from LiDAR point clouds and GNSS, forming a layer comprised of 512 units. Subsequently, these features are fused via two fully connected networks with 256 units respectively. The fused features inherit the presentation from the two perception inputs. These features leverage speed predicted branch \cite{speedbranch} to emphasize the utilization of localization information to extract cues that reflect the dynamics of the scenario. In addition, with the assistance of HLC, these features enable the prediction of reasonable decoupled control commands for autonomous mining trucks.

\subsection{Task Dependent Control}\label{sec:Task-dependentLos}

\textcolor{black}{Lateral and longitudinal control are two dependent tasks for mining trucks. On the one hand, the accomplishment of each task is dependent on the differential of scenario features, for example, the control schemes are totally different in transportation roads and dumping sites. Lateral control heavily relies on traffic signs and road structures, whereas longitudinal control is profoundly influenced by other transportation participants ahead and the speed of the ego-truck. On the other hand, these two control tasks show different tolerances towards perturbations in two control commands. When faced with identical scenarios, the confidence levels pertaining to lateral and longitudinal controls diverge, thereby emphasizing the inherent uncertainties associated with the two tasks.}

Within the field of multi-task learning, different deep networks are computed from each task, and separate learning models are integrated into a unified loss function \cite{MT}. A linear combination is a typical method that incorporates the weighing of individual task losses using hand-tuned hyperparameters \cite{MT1}. However, tuning for optimal hyperparameters is a hard task. As the performance of the trained model is frequently sensitive to these hyperparameters, its generalization may be limited in diverse scenarios, especially in open-pit mines with highly extreme weather. In alignment with \cite{MT2}, this research formulates the learning task of lateral and longitudinal control within a multi-task learning framework simultaneously. Specifically, task-dependent uncertainties are employed to predict the weights for each task. Moreover, these uncertainties are acquired from input data and jointly optimized alongside the model parameters. The two control tasks are recognized as two regression progresses in FusionPlanner that need to be trained separately. \textcolor{black}{$\pi_{\theta }\left ( s \right )$ presents a neural network policy parameterized by $\theta$ and also stands for the abbreviation of $\pi(o,c;\theta)$, with the state $s$ as input data and the control command $a$ as output data. The likelihood of this network is built up as a Gaussian distribution with the mean given by the model output, and the scalar $\sigma^{2}$ stands for the noise for task-dependent uncertainty:}

\begin{eqnarray}
    p(a|\pi_{\theta }(s)) = \mathcal{N}(\pi_{\theta}(s), \sigma^{2}) 
    \label{equ:2}
\end{eqnarray}

\begin{eqnarray}
    -\log{p(a|\pi_{\theta}(s))}  \propto \frac{1}{2\sigma^{2}}\left \| a-\pi_{\theta (s)} \right \|^2 + \log{\alpha}    
    \label{equ:3}
\end{eqnarray}

Considering the multi-task control task that computes two outputs, including a lateral signal with steering angle $a_{lat}$ and a longitudinal signal with acceleration $a_{lon}$. Assuming the two tasks are approximately independent, the following formula can be derived:
\begin{eqnarray}
    \begin{aligned}
            p(a_{lat},a_{lon}|\pi_{\theta }(s))&=p(a_{lat}|\pi_{\theta }(s))\cdot p(a_{lon}|\pi_{\theta }(s))\\ &= \mathcal{N}(a_{lat};\pi_{\theta },\sigma_{lat}^{2}) \cdot \mathcal{N}(a_{lon};\pi_{\theta },\sigma_{lon}^{2}) 
    \end{aligned}
    \label{equ: 4}
\end{eqnarray}

\begin{eqnarray}
    \begin{aligned}
            -\log {p(a_{lat},a_{lon}|\pi_{\theta }(s))}\propto \frac{1}{2\sigma _{lat}^{2}} \left \| a_{lat}-\pi^{lat} _{\theta }(s) \right \|^2 + \frac{1}{2\sigma _{lon}^{2}} \left \| a_{lon}-\pi^{lon} _{\theta }(s) \right \|^2 + \log{\sigma _{lat}\sigma _{lon}}
    \end{aligned}
    \label{equ: 5}
\end{eqnarray}

Consequently, the task-dependent uncertainly loss function $\mathcal{L}$ for multi-task learning of lateral and longitudinal controls can be formulated as:
\begin{equation}
    \mathcal{L}(\theta ,\sigma_{lat},\sigma_{lon})=  \frac{1}{2\sigma _{lat}^{2}} \left \| a_{lat}-\pi^{lat} _{\theta }(s) \right \|^2 + \frac{1}{2\sigma _{lon}^{2}} \left \| a_{lon}-\pi^{lon} _{\theta }(s) \right \|^2 + \log {\sigma _{lat}\sigma _{lon}}
    \label{equ:6}
\end{equation}
where lateral and longitudinal conditional control can be shown as $\pi_{\theta}^{lat}$ and $\pi_{\theta}^{lon}$. $\sigma_{lat}$ and $\sigma_{lon}$ represent the task-specific uncertainties associated with lateral and longitudinal controls. In the loss function $\mathcal{L}$, the first and second terms cover the objectives pertaining to each individual task, which are weighted by $\sigma_{lat}$ and $\sigma_{lon}$. Minimizing $\mathcal{L}$ with respect to $\sigma_{lat}$ and $\sigma_{lon}$ can be trained from input data. For instance, the lateral control task can be inherently inferred with more uncertainty due to the large value of $\sigma_{lat}$, then the task needs a smaller weight, and vice versa.

In most of the previous literature, the weights corresponding to lateral and longitudinal control losses are manually hyperparameterized by human craft \cite{CIL, Liu, CDQN}, this research adopts the method \cite{Zhu} that adaptively learns to dynamically balance these weights. The terminal term, $\log{\sigma_{lat}\sigma_{lon}}$, serves as a regularization mechanism to preventing $\sigma_{lat}$ and $\sigma_{lon}$ from excessively increasing.

 The mining truck is a specialized machinery designed for heavy transportation in open-pit mines. Its braking system differs significantly from that of common vehicles. The truck is equipped with two complementary braking systems: the first one is the electromagnetic retarder, also known as the electric brake, which serves as an auxiliary braking device for the mining truck and converts the kinetic energy of the truck into electrical energy through electromagnetic induction during braking, resulting achieving the deceleration function. The advantages of the electric brake include mechanical wear-free operation, smooth braking, and absence of impact and noise, etc. However, it is not capable of completely stopping the truck. Generally, this brake is used when the mining truck is being driven at high speeds. The second braking system is the mechanical brake, which relies on mechanical friction to stop the truck. This brake is typically employed to completely hold up the mining truck after the electric brakes have controlled the truck's speed to 5 km/h. For the mining truck, the longitudinal control can be divided into three parts: throttle \textit{$a_{acc}$}, electric brake \textit{$a_{dec\_e}$}, and mechanical brake \textit{$a_{dec\_m}$}. Thus, the Equ.\ref{equ:6} should be customized as follows:

 \begin{eqnarray}
    \begin{aligned}
        \mathcal{L}(\theta ,\sigma _{str}, \sigma_{acc}, \sigma_{dec\_e}, \sigma_{dec\_m}) &= \frac{1}{4\sigma _{str}^{2}} \left \| a_{str}-\pi^{str} _{\theta }(s) \right \|^2 + \frac{1}{4\sigma _{acc}^{2}} \left \| a_{acc}-\pi^{acc} _{\theta }(s) \right \|^2 + \frac{1}{4\sigma _{dec\_e}^{2}} \left \| a_{dec\_e}-\pi^{dec\_e} _{\theta }(s) \right \|^2 \\&+ \frac{1}{4\sigma _{dec\_m}^{2}} \left \| a_{dec\_m}-\pi^{dec\_m} _{\theta }(s) \right \|^2 + \log{(\sigma _{str} \sigma_{acc} \sigma_{dec\_e}\sigma_{dec\_m})},
    \end{aligned}
     \label{equ:7}
 \end{eqnarray}
where $a _{str}$ is the steering angle in lateral control, $a_{acc},a_{dec\_e},a_{dec\_m}$ demonstrate throttle, electric brake, and mechanical brake in longitudinal control. $\pi_{\theta}(s)$ shows the same meaning as Equ.\ref{equ:6}, but it is composed of four contents. $\sigma _{str}, \sigma_{acc}, \sigma_{dec\_e}, \sigma_{dec\_m}$ represent the weight for the task-specific uncertainties associated with four control commands. The final term, $\log{(\sigma _{str} \sigma_{acc} \sigma_{dec\_e}\sigma_{dec\_m})}$, also serves as a regularization term to preventing $\sigma_{str}$,$\sigma_{acc}$,$\sigma_{dec\_e}$ and $\sigma_{dec\_m}$ from excessively changing.

\textcolor{black}{In addition, as depicted in Fig.\ref{fig:framework}, Rule-based Behaviour Decision module offers high-level commands for FusionPlanner. These commands determine the branch selection for both lateral and longitudinal control, taking into account variations in velocity and steering. The lateral commands include straightforward, left turns, and right turns, while the longitudinal commands cover deceleration, acceleration, and maintaining current speed.}

\begin{figure}[H]
    \centering
    \includegraphics[width=0.85\linewidth]{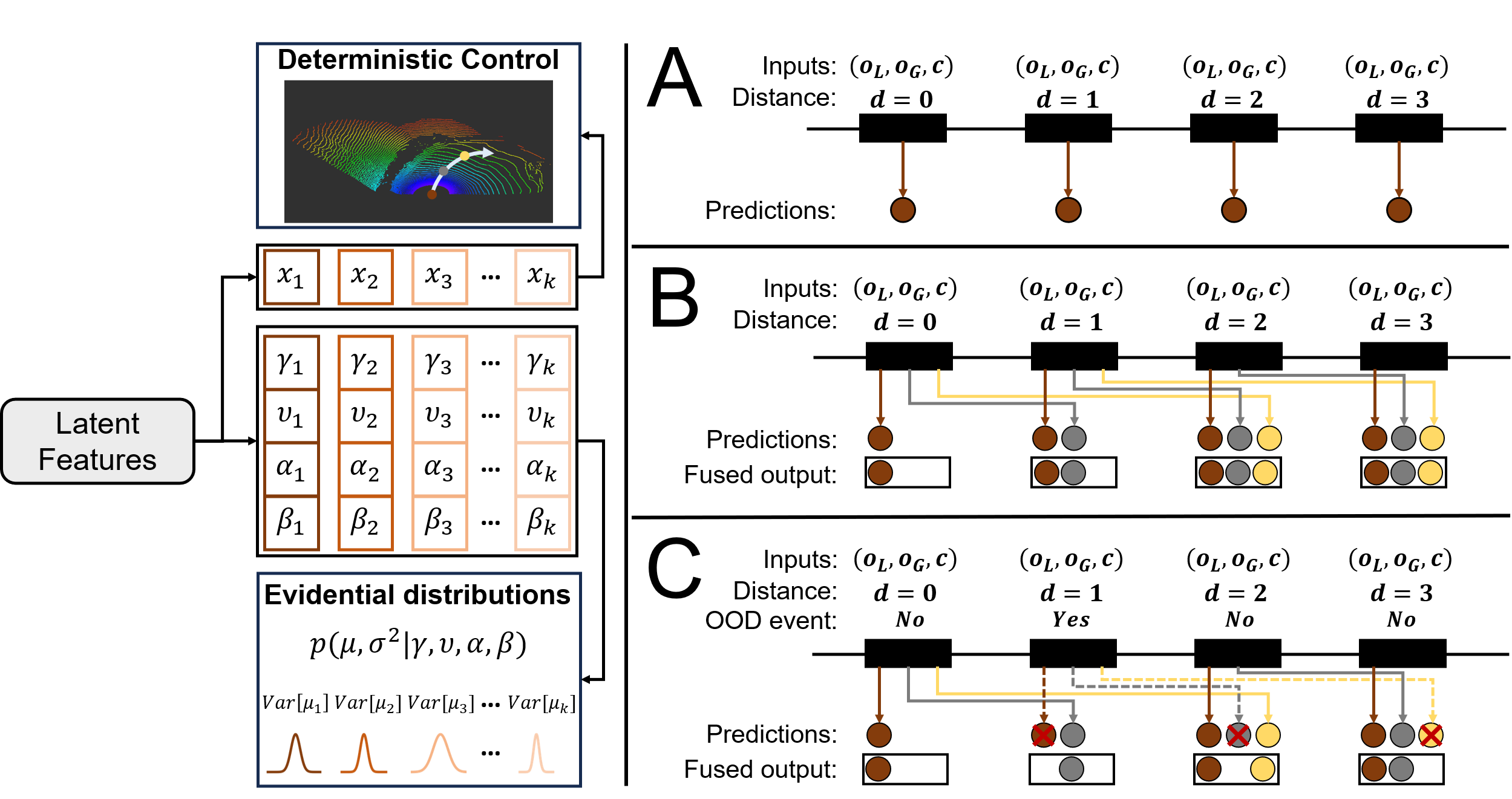}
    \caption{The output control commands of FusionPlanner can be implemented in three modes. \textbf{(A)} Instantaneous model: executing all commands instantaneously; \textbf{(B)} Uniform fusion model: uniformly fuse the predictions from multiple past frames; \textbf{(C)} Evidential fusion model: incorporating uncertainty estimation in the output, intelligently weighting the predicted commands to enhance the trustworthiness and robustness, particularly in OOD events or on increased uncertainty of future time steps.}
    \label{fig:evidentialfusion}
\end{figure}

\subsection{Hybrid Evidential Fusion}\label{sec:evidentialfusion}

FusionPlanner, as shown in Fig.\ref{fig:framework}, directly computes control commands $\left \{ a_{k} \right \} $ for the autonomous mining truck, which can be supervised with the labeled control commands $\left \{ y_{k} \right \}$ with L1 Loss: $\mathcal{L}_{MAE} (a_{k},y_{k})=\left \| a_{k}-y_{k} \right \|_{1}$. \textcolor{black}{The predicted commands can be directly implemented in the autonomous truck, as shown in Fig.\ref{fig:evidentialfusion} \textbf{(A)}.} However, open-pit mines are always characterized by complex operational conditions and adverse environmental factors. Severe weather conditions such as blizzards, rainstorms, and gale-force winds, as well as sensor failures, may lead to unexpected control commands by the end-to-end planner.

To alleviate this problem, our model tries to compute additional knowledge concerning the epistemic uncertainty associated with each predicted command, rather than rigidly enforcing the execution of every prediction, as shown in Fig.\ref{fig:evidentialfusion} \textbf{(A)}. To achieve this, we incorporate an auxiliary branch called the hybrid evidential fusion model, which generates an evidence distribution for each prediction output. The proposed evidence distribution captures the confidence level associated with the selected predictions made by FusionPlanner, as shown in Fig.\ref{fig:evidentialfusion} \textbf{(C)}. This allows us to properly compare predictions with lower confidence to those with higher confidence within the network, facilitating the selection of high-quality predictions. In addition, the integration of control commands at each frame serves as a rational scheme for stabilizing control in Fig.\ref{fig:evidentialfusion} \textbf{(B)}.

Assume that the expert label control commands $y_{k}$ adhere to the Gaussian distribution with unidentified mean and variance $(\mu,\sigma^{2})$, and we explore its optimal distribution. In the context of learning objectives, such as the optimal control problem of mining trucks, we combine prior knowledge with likelihood variables to calculate a joint distribution with four unknown parameters, $p(\mu,\sigma^{2}|\gamma, \upsilon, \alpha, \beta )$ with

\begin{eqnarray}
    \begin{aligned}
        \mu\sim \mathcal{N}(\gamma, \sigma^{2} \upsilon^{-1} ), \\ \sigma^{2}\sim \Gamma^{-1}(\alpha ,\beta ),    
    \end{aligned}
\end{eqnarray}

FusionPlanner is trained to learn this unknown Gaussian distribution, $e_{k}=(\gamma_{k}, \upsilon_{k}, \alpha_{k}, \beta_{k})$, by jointly maximizing the loss ($\mathcal{L}_{NLL} $) on model fit and minimizing the loss ($\mathcal{L}_{R} $) for evidence on errors:

\begin{eqnarray}
    \mathcal{L}_{NLL}(w_{k},y_{k})=\frac{1}{2} \log{\frac{\pi}{\upsilon_{k}}}-\alpha _{k}\log{(2\beta_{k}(1+\upsilon_{k}) )} +(\alpha _{k}+\frac{1}{2} )\log{[(y_{k}-\gamma_{k})^2\upsilon_{k}+2\beta_{k}(1+\upsilon_{k})]}+\log{(\frac{\Gamma(\alpha_{k})}{\Gamma(\alpha_{k}+\frac{1}{2} )}) }
\end{eqnarray}

\begin{equation}
    \mathcal{L}_{R}(w_{k},y_{k})=\left | y_{k}-\gamma_{k}  \right | \cdot (2\alpha_{k}+\upsilon_{k})
\end{equation}

For further details regarding epistemic uncertainty and evidential distributions, please refer to \cite{deregression}. Finally, the total loss for our end-to-end planner coupled with hybrid evidential fusion can be formulated as follow:

\begin{equation}
    \mathcal{L}(\cdot )=\sum_{k}(\alpha \mathcal{L}_{MAE}(\cdot)+\mathcal{L}_{NLL}(\cdot )+\mathcal{L}_{R}(\cdot )) 
    \label{equ: total}
\end{equation}
where $\alpha$ represents a large scalar, 1500 in FusionPlanner, to balance the scale difference. The epistemic uncertainty can be inferences as $Var[x_{k}]={\beta_{k}}/[\upsilon_{k}(\alpha_{k}-1)]$. During deployment, the deterministic prediction $x_{k}$ as well as the corresponding evidential uncertainty $Var[x_{k}]$ can be collected from former frames. and then the uncertainty is utilized to evaluate a confidence-weighted average of predicted commands from our end-to-end planner.

\renewcommand{\algorithmicrequire}{\textbf{Input:}}
\renewcommand{\algorithmicensure}{\textbf{Output:}}
\algnewcommand\algorithmicswitch{\textbf{switch}}
\algnewcommand\algorithmiccase{\textbf{case}}
\algnewcommand\algorithmicassert{\texttt{assert}}
\algnewcommand\Assert[1]{\State \algorithmicassert(#1)}%
\algdef{SE}[SWITCH]{Switch}{EndSwitch}[1]{\algorithmicswitch\ #1\ \algorithmicdo}{\algorithmicend\ \algorithmicswitch}%
\algdef{SE}[CASE]{Case}{EndCase}[1]{\algorithmiccase\ #1}{\algorithmicend\ \algorithmiccase}%
\algtext*{EndSwitch}%
\algtext*{EndCase}%

\begin{algorithm}[h] \footnotesize
  \caption{Uncertainty-Aware Deployment}
  \label{alg1}
  \begin{algorithmic}[1]
    \Require
      policy $\pi_{\theta}$, inputs $(o_{L},o_{G},c)$, and fusion method $\left \{ \left ( a_{k},\gamma _{k},\upsilon _{k},\beta  \right )  \right \} \leftarrow \pi_{\theta}(o_{L},o_{G},c)$ in locally distance $d$, shown in Section \ref{sec:evidentialfusion}
 
    \For {$k\in \left \{ 0,1,2,...,K-1 \right \}$}          
        \State $Var[\mu _{k}]\gets \beta _{k}/(\upsilon _{k}(\alpha _{k}-1))$ \Comment{Compute uncertainty}
        \State $\lambda_{k} \gets   1/Var[\mu _{k}]$    \Comment{Compute confidence}
        \State $\Lambda^{d+k}\gets \Lambda^{d+k}\cup  \left \{ \lambda_{k} \right \} $ \Comment{Store confidence}
        \State $\chi ^{(d+k)}\gets  \chi ^{(d+k)} \cup\left \{ x_{k} \right \} $ \Comment{Store prediction}
    \EndFor

    \State $\Lambda^{d} \gets \Lambda^{d}/\sum_{\lambda  \in \Lambda^{d}}\lambda $ \Comment{Normalization}

    \Switch{fusion}
        \Case{none}
            \State \textbf{return:} $a_{0}^{(d)}$ \Comment{Instantaneous shown in Fig.\ref{fig:evidentialfusion} \textbf{(A)}}
        \EndCase
        \Case{uniform}
            \State \textbf{return:} $(\sum_{j}\chi _{j}^{(d)}) / \left \| \chi^{(d)} \right \|$  \Comment{Uniform fusion shown in Fig.\ref{fig:evidentialfusion} \textbf{(B)}}
        \EndCase
        \Case{evidential} 
            \State \textbf{return:} $(\sum_{j}\chi _{j}^{(d)} \Lambda_{j}^{(d)}) / \left \| \chi^{(d)} \right \| $ \Comment{Evidential fusion shown in Fig.\ref{fig:evidentialfusion} \textbf{(C)}}
        \EndCase
    \EndSwitch
        
  \end{algorithmic}
\end{algorithm}





\section{MiningNav Benchmark}
Building upon the distinctive environment of open-pit mines and the complex requirements of mineral transportation, we propose the first testing benchmark, named MiningNav, specifically designed for validating autonomous mining trucks on transportation roads. Considering the differences in production transportation within open-pit mines compared to the previous benchmark \cite{carla, Zhu, offline, prakash2021multi, chen2022learning}, MiningNav focuses on testing the trustworthiness of autonomous trucks throughout the transportation process. Specifically, we build our benchmark based on the intricate and authentic digital open-pit mining environments provided by PMS, and we design three verification experiments aimed at validating the robustness of the algorithm.

\subsection{Evaluation Metrics}

In the MiningNav benchmark, we design three specific experiments to validate the trustworthiness and robustness of the algorithm in the open-pit mines, including \textit{lane-stable task}, \textit{disturbance task}, and \textit{navigation task}. The lane-stable task focuses on evaluating the algorithm's robustness on conventional transportation roads. We select relatively smooth and bi-directional roads with a minimum length of 1000 m. The purpose is to evaluate how well the algorithm performs in maintaining a stable lane position throughout the given trajectory. The disturbance task imposts perturbations to the initial state of the autonomous mining trucks, including bias in heading angle and lateral deviation. The trucks are required to autonomously return to a safe state within a specified timeframe. This task serves to evaluate the algorithm's ability to handle OOD events and recover to a stable state. The navigation task involves initializing autonomous mining trucks at random starting points within the open-pit mine scenario. The objective of this task is to reach a destination that is at least 1000 meters away from the starting point and includes at least one intersection. This task is utilized to evaluate the algorithm's capability in navigating through complex scenarios and crossing unstructured intersections.

In the closed-loop simulation used for validation, the current raw perception data and high-level commands are fed into the motion planner. As the PSM only accepts control command values within the range of [-1:0; 1:0], the control command from the planner is mapped to this range and forwarded to the controller within PSM. The backend engine simulates the entire inference of the open-pit mine and proceeds to the next simulation step. This process is repeated until a task is finished. Since expert human drivers may make different control decisions in similar situations, randomness is introduced in command selection. Therefore, there is no single standard answer for validating the planner. \textcolor{black}{Consequently, the MiningNav benchmark employs different metrics that cover potential OOD events occurring in each episode to validate the trustworthiness and robustness of the well-trained planner in different tasks. 1). In the lane-stable task, we employ the number of collisions/interventions to evaluate the robustness of the algorithm in a selected transport path, which provides insights into the algorithm's ability to navigate the chosen path effectively. 2). In the disturbance task, our criterion is for the autonomous mining truck to revert to a safe state within a predetermined timeframe. Episodes that exceed the threshold or collisions are deemed as failures, which evaluate the trustworthiness of the algorithm in corner cases. 3). In the navigation task, our primary objective is to verify the algorithm's ability to through intersections within the transportation network of the open-pit mine. The autonomous truck is passed successfully if it completes the entire steering process without any intervention, which evaluates the algorithm's competence in navigating the complex network of transportation roads.}




\subsection{Dataset Collection}
The MiningNav dataset consists primarily of two components totaling 15 hours. The first component, which accounts for approximately 40\% of the dataset, involves the collection of driving demonstrations performed by skilled experts. The skilled experts are equipped with real-time front-view RGB images and BEV (Bird's Eye View) maps in the collection process. The BEV maps show reference lines and surrounding environments, providing task-related cues for the experts. Throughout the data collection process, the experts are instructed to adhere to transportation regulations, ensuring that the mining trucks maintain a speed limit of 20 km/h and follow HLC in intersections. The second component constitutes approximately 60\% of the dataset and is obtained through the integration of the autonomous driving model with the PMS simulator. This component captures the corresponding perception information and control commands generated by the autonomous driving model. Its purpose is to ensure a substantial volume of data, providing comprehensive coverage of various scenarios.

\section{Parallel Mine Simulator}
In comparison with the application of autonomous driving on urban roads, the complex conditions, harsh environments, rough roads, and frequent accidents in open-pit mines heighten the urgency for autonomous driving. However, conducting direct training and testing in mining scenarios poses significant challenges due to the complex experimental conditions and high associated costs. To address these challenges, we develop a high-fidelity simulation platform named Parallel Mine Simulator (PMS). This simulator showcases several highly realistic open-pit mining environments, providing researchers with a platform to collect and analyze manufacturing data, as well as test and compare various algorithms. The primary objective of PMS is to facilitate advancements in unmanned transportation within the mining industry and make significant contributions toward achieving autonomous mining operations.


\begin{figure}[H]
\centering  
\subfigure[Transport sections]{
\includegraphics[width=0.4\textwidth,height=0.22\textwidth]{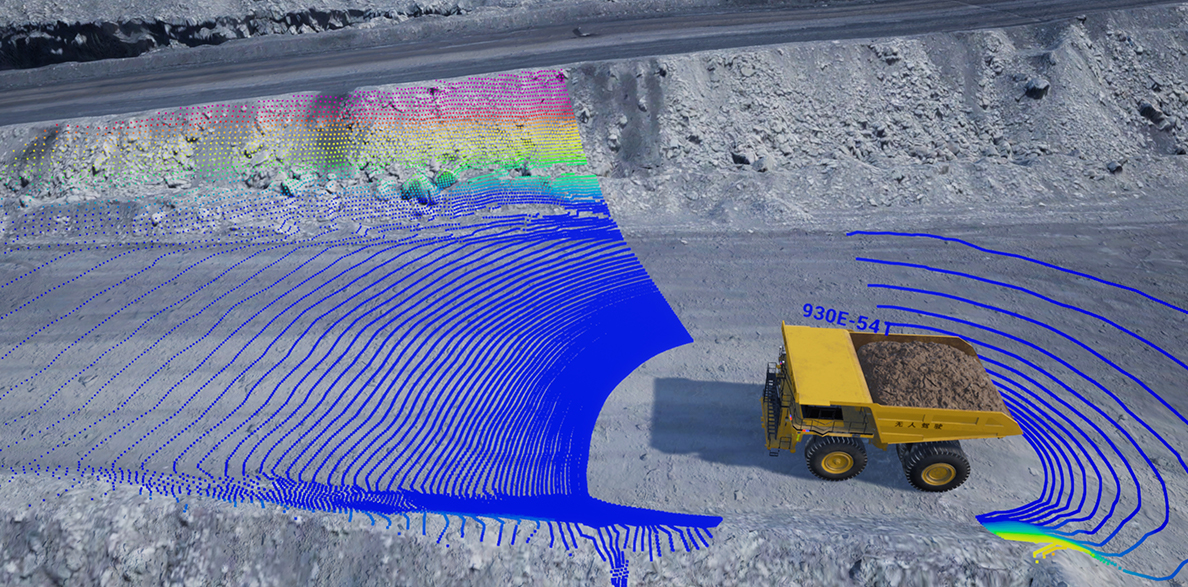}
\label{fig:simulator}
}
\subfigure[Production operation scene]{
\includegraphics[width=0.4\textwidth,height=0.22\textwidth]{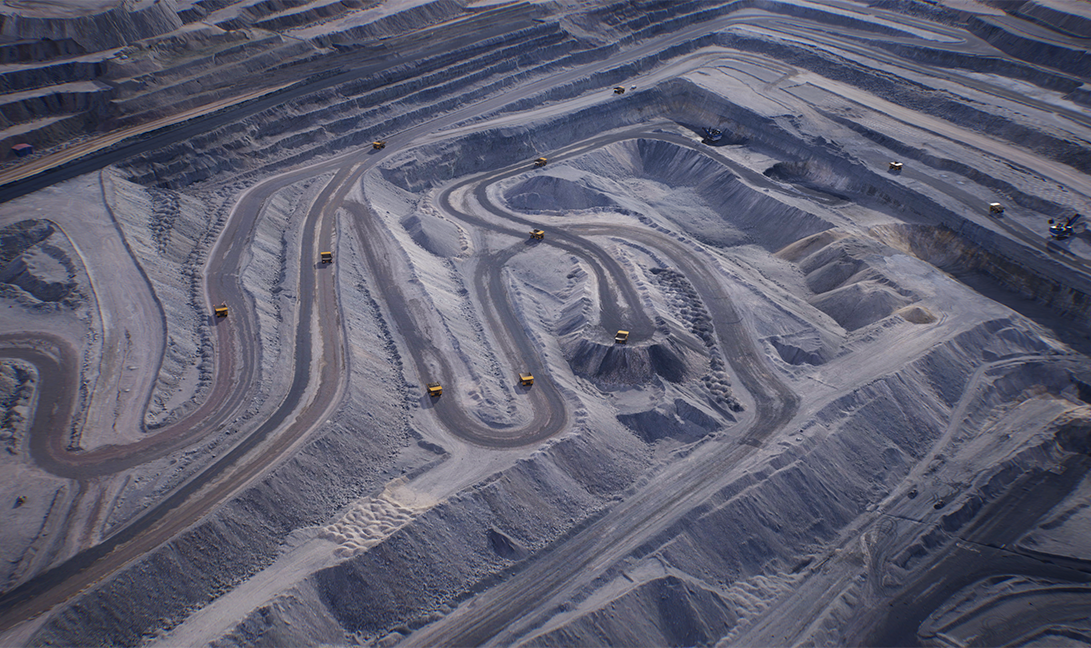}
\label{fig:env}
}
\caption{Part of scenarios of the open-pit mine in PMS.}
\end{figure}

To the best of our knowledge, PMS stands as the pioneering open-pit mining simulation platform specifically designed for unmanned transportation, encompassing numerous distinctive characteristics inherent to open-pit mines. As illustrated in Fig.\ref{fig:simulator}, open-pit mines are prototypical unstructured scenarios, which introduce significant challenges in the extraction of scenario features due to the absence of texture and edges in the surrounding. This poses a major challenge for various autonomous driving algorithms, that achieve excellent performance in structured scenarios with traffic lanes and distinct feature information, but are prone to failure in open-pit mines. Furthermore, in comparison with urban scenarios, the participants involved in open-pit mine transport are of gigantic larger size often manifesting instances of partial perception, thereby presenting monumental challenges to classification and localization. Consequently, the detection of potential risk in unstructured scenarios with limited feature richness and the augmentation of the reliability and robustness of autonomous driving algorithms constitute unique contributions that our simulator offers to researchers.

\subsection{Simulator Functions}
PMS serves as a comprehensive simulator for the total transportation process within open-pit mines and provides a multitude of control interfaces between the simulator and the transportation participants. To fulfill this purpose, PMS is architected as a traditional one-server, multi-client framework. The server generates a rendered environment of the open-pit mine and programmatic transportation participants, while the clients, acting as autonomous agents, establish communication with the server through a sophisticated transmission system. The server is capable of accommodating multiple clients simultaneously. The clients receive raw perception data from the server as inputs to the autonomous driving framework, subsequently dispatching control commands to the server to manage the agent's behavior, including steering, acceleration, and braking. Moreover, PMS provides flexibility for server parameter adjustments, such as simulator resets, modifications of environmental attributes, and alterations of sensor configurations. Given the harsh environmental factors in open-pit mines, PMS presents a wealth of environmental attributes, including meteorological conditions, luminosity intensity, precipitation density, dust particulate density, and more.

\subsubsection{Environments}
The rendered environment in the server is composed of 3D models of static elements, including buildings, traffic signs, and infrastructures, as well as production machines such as mining trucks, bulldozers, and excavators, and auxiliary vehicles such as water sprinklers and bulldozers. Each element is fine-tuned to ensure visual fidelity, the geometric models and textures are used to ensure a visually realistic experience within PMS. To maintain authenticity, all elements adhere to a consistent scale, accurately reflecting their real-world dimensions. Currently, our simulator comprises 30 distinct building models, 20 production machine models, and 5 auxiliary vehicle models. PMS constructs the open-pit mine using the assets through the following steps: 
\begin{itemize}
    \item[1)] Importing the production operation zones, including loading sites, transportation roads, and dumping sites;
    \item[2)] Incorporating relevant boundaries and textures of the entire open-pit mine;
    \item[3)] Manually placing buildings, terrains, and traffic infrastructures;
    \item[4)] Specifying the area where traffic participants can work.
\end{itemize}

As depicted in Fig.\ref{fig:env}, an open-pit mine is built within PMS, with each mine capable of accommodating up to 200 mining trucks operating simultaneously.

\begin{figure}[H]
\centering  
\subfigure[Clear Day]{
\label{Fig:weather.1}
\includegraphics[width=0.3\textwidth]{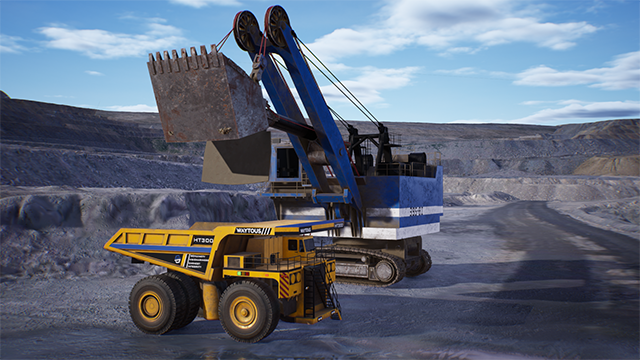}}
\subfigure[Daytime Rainy]{
\label{Fig:weather.2}
\includegraphics[width=0.3\textwidth]{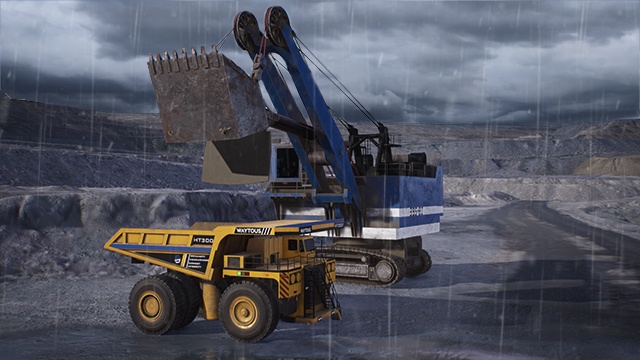}}
\subfigure[Daytime Sandy]{
\label{Fig:weather.3}
\includegraphics[width=0.3\textwidth]{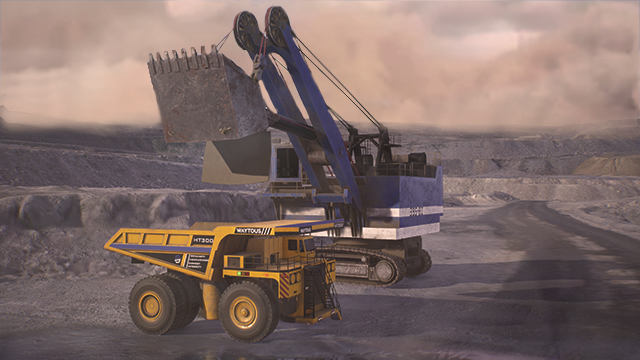}}
\caption{The loading site in PMS, shown from the third-person view in three common weather conditions of open-pit mines.}
\label{Fig.weather}
\end{figure}

\subsubsection{Sensors}
\begin{figure}[H]
\centering  
\subfigure[Third-Person View]{
\label{Fig:ect.1}
\includegraphics[width=0.3\textwidth]{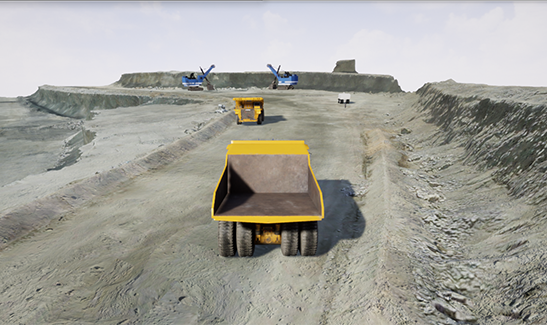}}
\subfigure[RGB Image]{
\label{Fig:ect.2}
\includegraphics[width=0.3\textwidth]{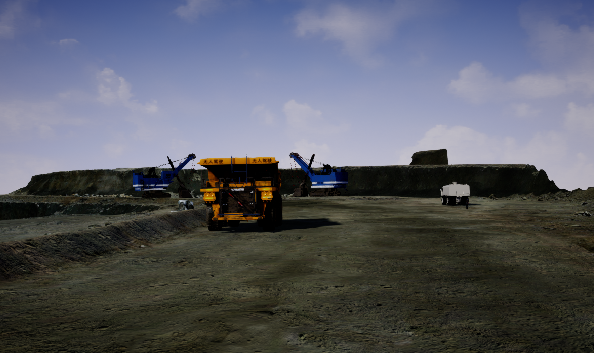}}
\subfigure[Semantic Segmentation Image]{
\label{Fig:ect.3}
\includegraphics[width=0.3\textwidth]{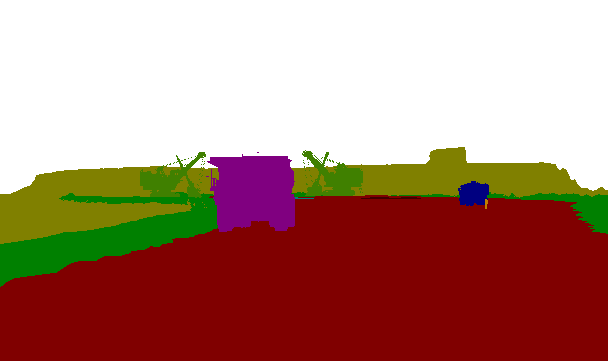}}
\caption{Three of the sensing modalities provided by PMS. Semantic segmentation image belongs to pseudo-sensors and is supported as a label for perception experiments. Additional sensor models can be mounted in the simulator.}
\label{Fig.Dect}
\end{figure}

Sensors play a pivotal role in the advancement of autonomous driving, and PMS provides flexible sensor configuration options for the agents. Currently, PSM implements various sensors designed for the operational requirements in the open-pit mine, including LiDAR,  MMV radar (millimeter-wave radar), and onboard cameras, as depicted in Fig.\ref{Fig.Dect}. The LiDAR system in PMS encompasses six different types of LiDAR, ranging from 1-line to 128-line, covering all mainstream LiDAR variants. The MMV radar primarily consists of long-range and short-range types, enabling effective object detection. Onboard cameras are equipped to capture a wealth of visual information, including RGB images, depth images, grayscale images, semantic images, and fisheye images. To enhance the role of vision in open-pit mining scenarios, semantic segmentation in PMS provides 24 semantic categories for the cameras, such as sky, drivable areas, retaining walls, mining trucks, excavators, passenger vehicles, water sprinklers, bulldozers, power shovels, specialized engineering machinery, fences, utility poles, large rocks, and so on. Considering the extreme weather conditions often encountered in open-pit mines, PMS offers diverse configurations related to weather conditions, as shown in Fig.\ref{Fig.weather}. This allows for the validation of model trustworthiness and robustness by introducing environmental perturbations. By validating the impact of these perturbations on sensor accuracy, the evaluation of algorithmic robustness is facilitated.

\subsubsection{Transportation Participants}

In addition to the raw perception information from various sensors, PMS provides a comprehensive set of vectors related to the agent's states, which include local coordinates and GNSS coordinates, velocity, acceleration vectors, angular velocity, and angular acceleration. Furthermore, PSM also presents some measurements related to transportation regulations, including the percentage of vehicles deviating from the reference line, as well as the current road conditions and speed limits at the current position. Additionally, PMS furnishes precise location and bounding box outputs for all traffic participants, which are crucial for training and validating autonomous driving algorithms.

In contrast to the diverse traffic participants in urban roadways, the participants within open-pit mines exhibit a more concentrated distribution of categories. These categories primarily consist of production vehicles such as mining trucks, bulldozers, and excavators, as well as auxiliary vehicles such as water sprinklers and bulldozers. Among these categories, mining trucks hold particular significance for mineral transportation. Therefore, PMS incorporates high-fidelity models for the current mainstream mining truck, such as 730E, 830E, and 930E from Komatsu, 797F from Caterpillar, EH5000 from Hitachi. Additionally, the model repository in PMS encompasses other transportation participants, allowing researchers to easily deploy a wide range of vehicles. All special machines within PMS are equipped with high-fidelity dynamics and kinematics models. The chassis, wheels, and joints are constructed with appropriate constraint conditions, ensuring more precise control feedback is reflected in the simulated world.

\section{Experiments}
Our motion planner is deployed in the mining truck equipped with an optimal controller that fulfills all the prerequisites required by the planner. In perception, the primary sensor of the truck is a Velodyne HDL-64E LiDAR, which has an output frequency of 10Hz. All computations related to FusionPlanner are carried out on an NVIDIA RTX 2080ti GPU. In localization, the mining truck can directly receive localization information directly from the embedded localization system within the PMS. 

\subsection{Data Processing}

After the data collection, we filter biased data and prepare for training. While expert drivers have maintained a high level of attentiveness during the data collection process, occasional oversights are unavoidable due to the extended duration of driving, such as too close to retaining walls and abrupt acceleration/deceleration \cite{xia2023automated}. The mining truck is equipped with four control commands (steering, throttle, electric brake, and mechanical brake). We specifically focus on steering and throttle for further supervising and filtering. Bias data are identified and removed if one of them is out of the predefined threshold ($i.e., a_{str}\notin (\in_{str}^{low},\in_{str}^{up}),a_{acc}\ge  (\in_{acc}^{up})$). 

\begin{figure}[H]
\centering  
\subfigure[Histogram of steering value]{
\label{Fig:steerdistribution}
\includegraphics[width=0.45\textwidth]{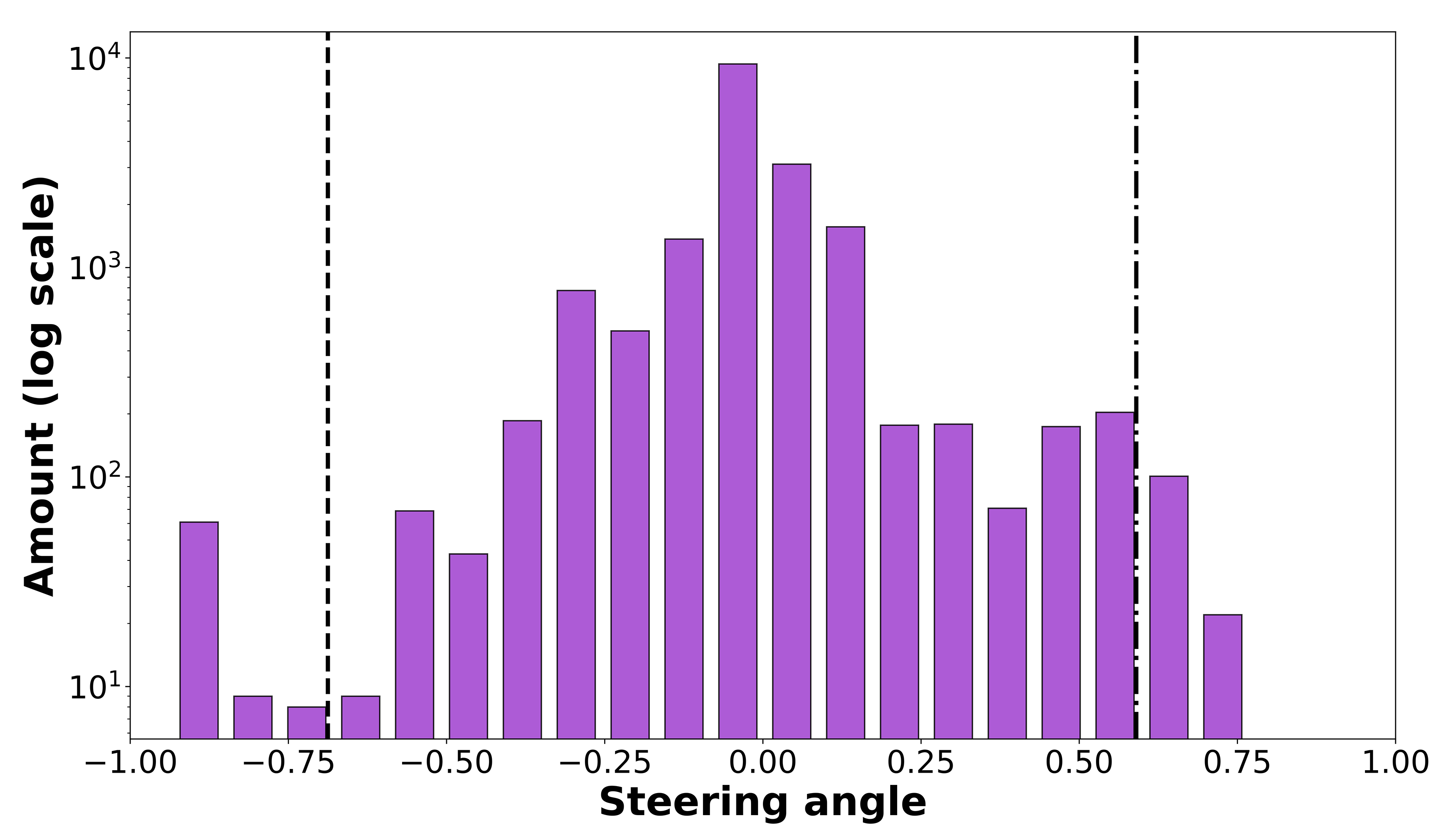}}
\subfigure[Histogram of throttle]{
\label{Fig:accdistribution}
\includegraphics[width=0.45\textwidth]{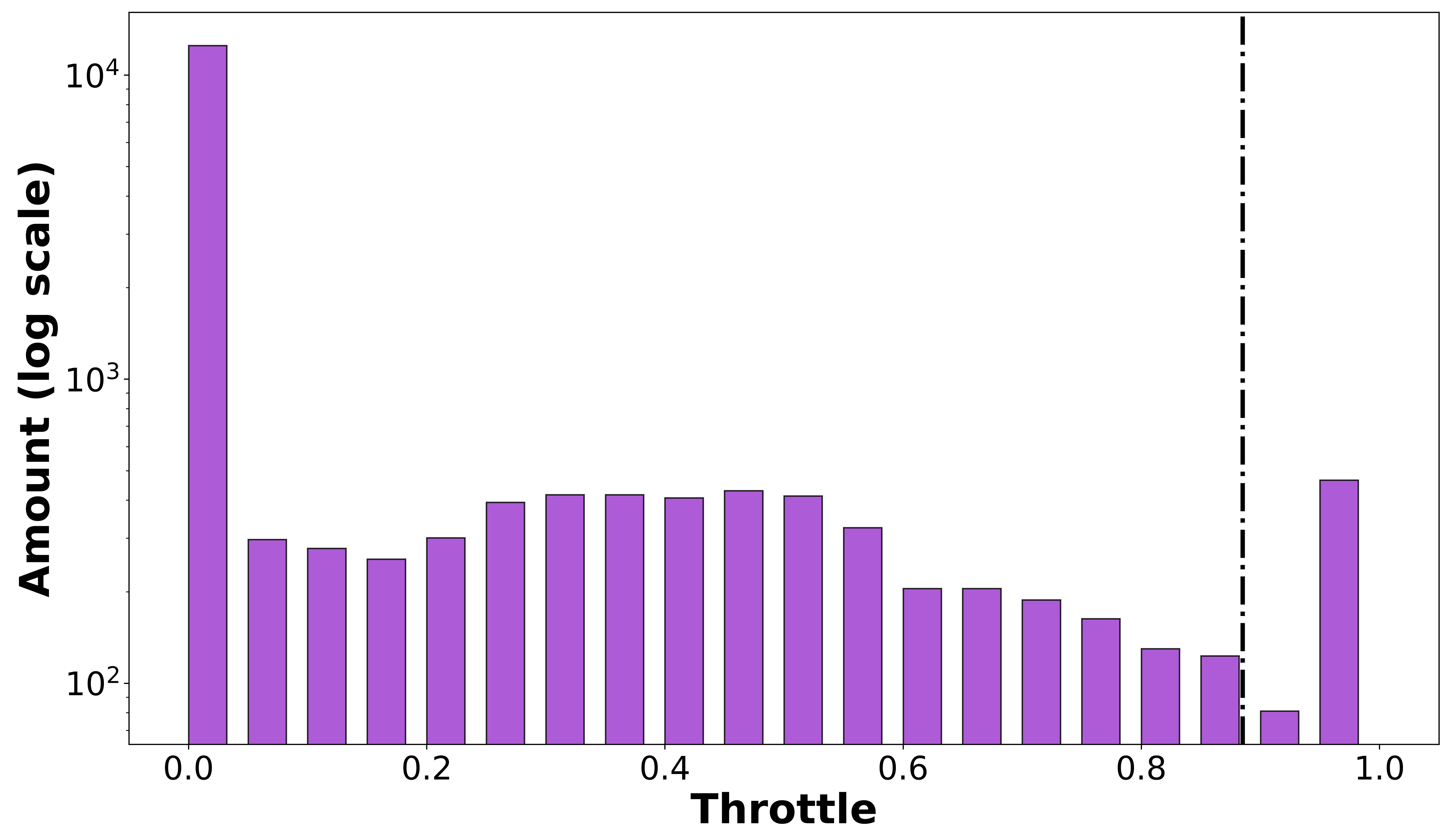}}
\caption{The steering angle and throttle are respectively mapped into [-1, +1] and [0, +1]. The left dashed line represents the lower bound of the confidence interval, while the right dashed line represents the upper bound. During the operation of the mining truck, the throttle remains unchanged for the majority of the time, only biased commands bigger than the upper bound are filtered for the throttle. (Note: y-axis is logarithmically scaled to represent the amount.)}
\label{Fig.hist}
\end{figure}

The thresholds $(\in_{str}^{low},\in_{str}^{up}, \in_{acc}^{up})$ mentioned above are determined based on an analysis of the collected control commands. Two histograms are generated to illustrate the distribution of the two control commands, as shown in Fig.\ref{Fig.hist}.To ensure a higher level of data quality and reduce the presence of biased operations, we leverage the upper and lower bounds of the confidence interval corresponding to a 0.99 confidence level. Any data points that fall outside of this confidence interval are considered biased and subsequently excluded. This process helps improve the overall quality of the dataset, enabling FusionPlanner to minimize the prediction of erroneous commands and enhance its robustness.

For each raw point cloud, we filter out all points within a distance of less than 4m or beyond 120m from the LiDAR and then quantize the coordinates of the remaining points with the voxel size of 0.2m, reducing the total point count to around 45,000. FusionPlanner predicts $k$ lookaheads control commands to address the stable control problem. Extracting the next two frames as references, we leverage a linear interpolation algorithm to obtain the four control commands for the $k$ lookaheads. These control commands serve as supervisors during the training process, guiding the network toward more accurate predictions.

\subsection{Data Argumentation}
Data augmentation can enable models to observe additional novel scenarios that are without the distribution of the current dataset, thereby enhancing the trustworthiness and robustness of the planner. The specific steps are implemented for point clouds: Firstly, we randomly scale the point cloud by a uniformly sampled coefficient from the interval [0:95; 1:05]. The corrections are also applied to the labeled control commands, keeping the truck on the correct trajectory. Subsequently, we randomly rotate the scaled point cloud by a small yaw angle $\theta \pm 10^{\circ}$. For the localization message, we introduce a subset that eliminates the GNSS, accounting for approximately 0.3\% of the total dataset, to reinforce the generalization of the motion planning model in OOD events.

\subsection{Model Training}
Frames with smaller absolute control commands contribute less to the model optimization, which can potentially lead to catastrophic results once the model predicts inaccurate controls, causing the truck to deviate from the road. To mitigate this, we multiply each term in Equ.\ref{equ: total} by a scalar factor $(1+exp[-y_{k}^{2}/(2\sigma^{2} )])$, which boosts the loss magnitude for frames with small control commands, thereby enhancing the learning efficiency of the planner. $\sigma$ is always equal to $1/15$ in all experiments. Our motion planner is trained for 250 epochs with 32 batch sizes. The selected optimizer is ADAM with $\beta_{1}=0.9$ and $\beta_{2}=0.999$, and the initial learning rate is $2e-4$ with a cosine decay schedule.

\section{Results}

\subsection{Lane-Stable Task}
\begin{figure}[t]
\centering  
\subfigure[Instantaneous with clockwise]{
\label{Fig:sub.1}
\includegraphics[width=0.2\textwidth]{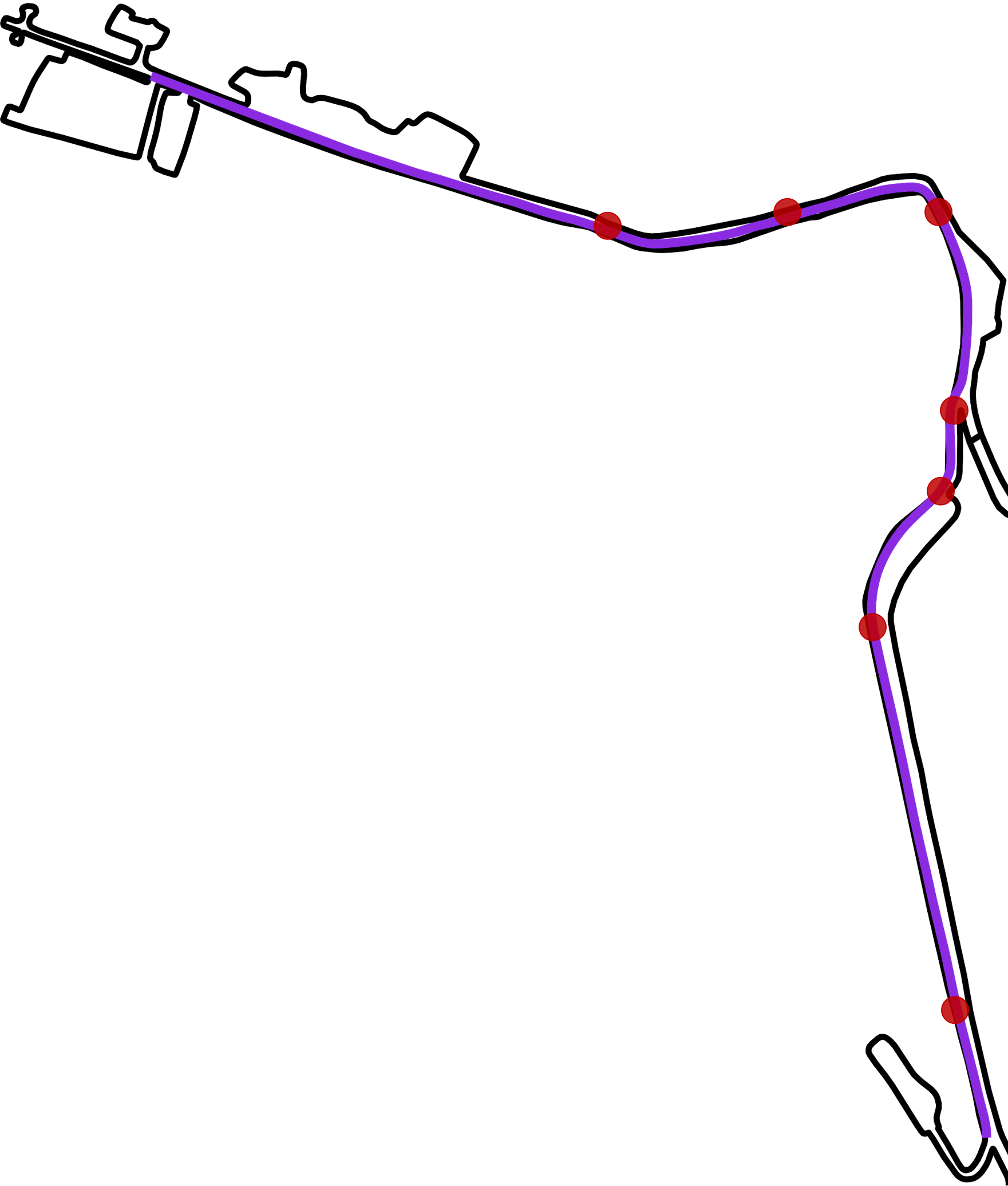}}
\subfigure[Uniform fusion with clockwise]{
\label{Fig:sub.2}
\includegraphics[width=0.2\textwidth]{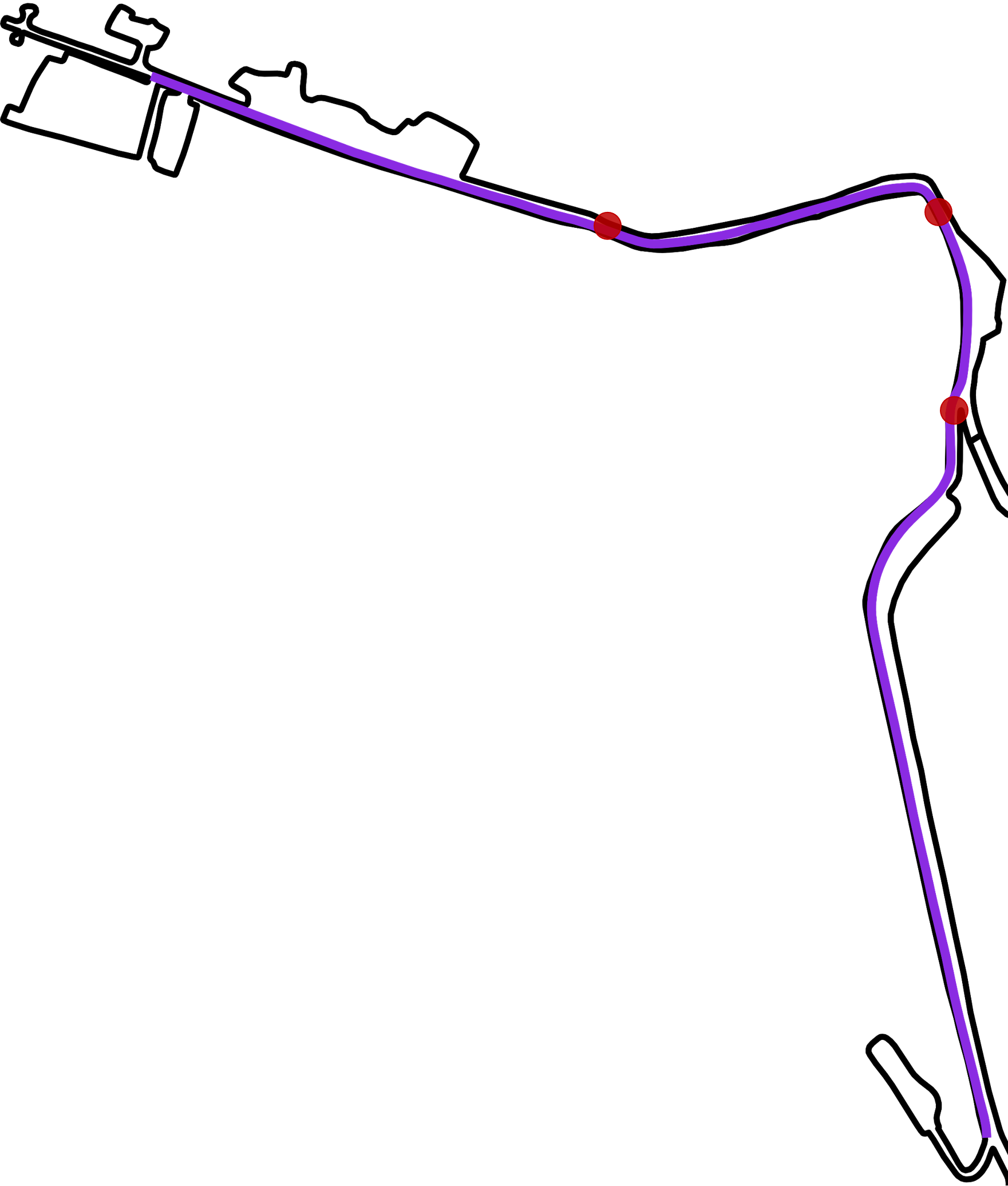}}
\subfigure[Evidential fusion with clockwise]{
\label{Fig:clocknormal}
\includegraphics[width=0.2\textwidth]{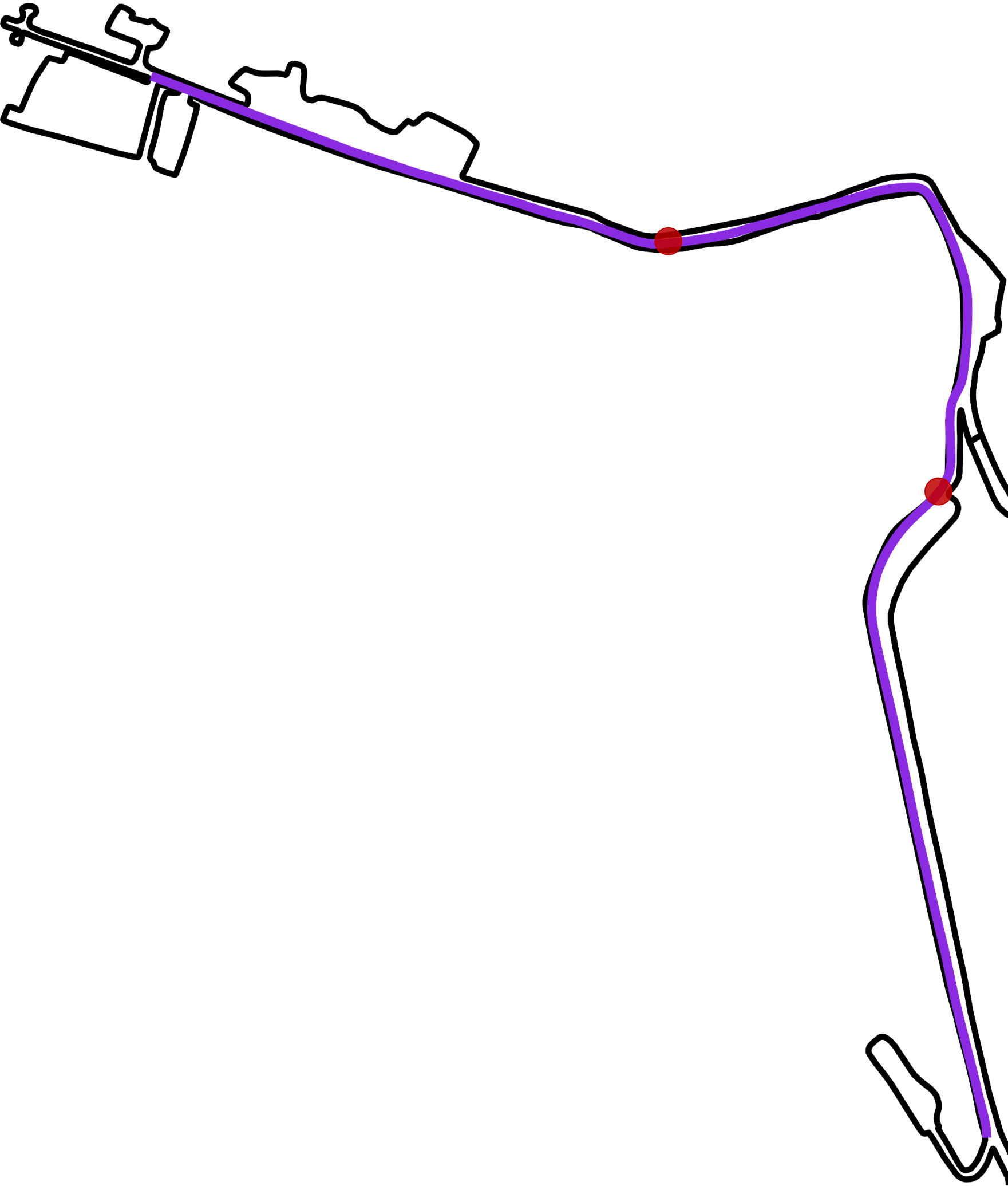}}
\subfigure[Evidential fusion with clockwise in OOD]{
\label{Fig:gpsOODclockwise}
\includegraphics[width=0.2\textwidth]{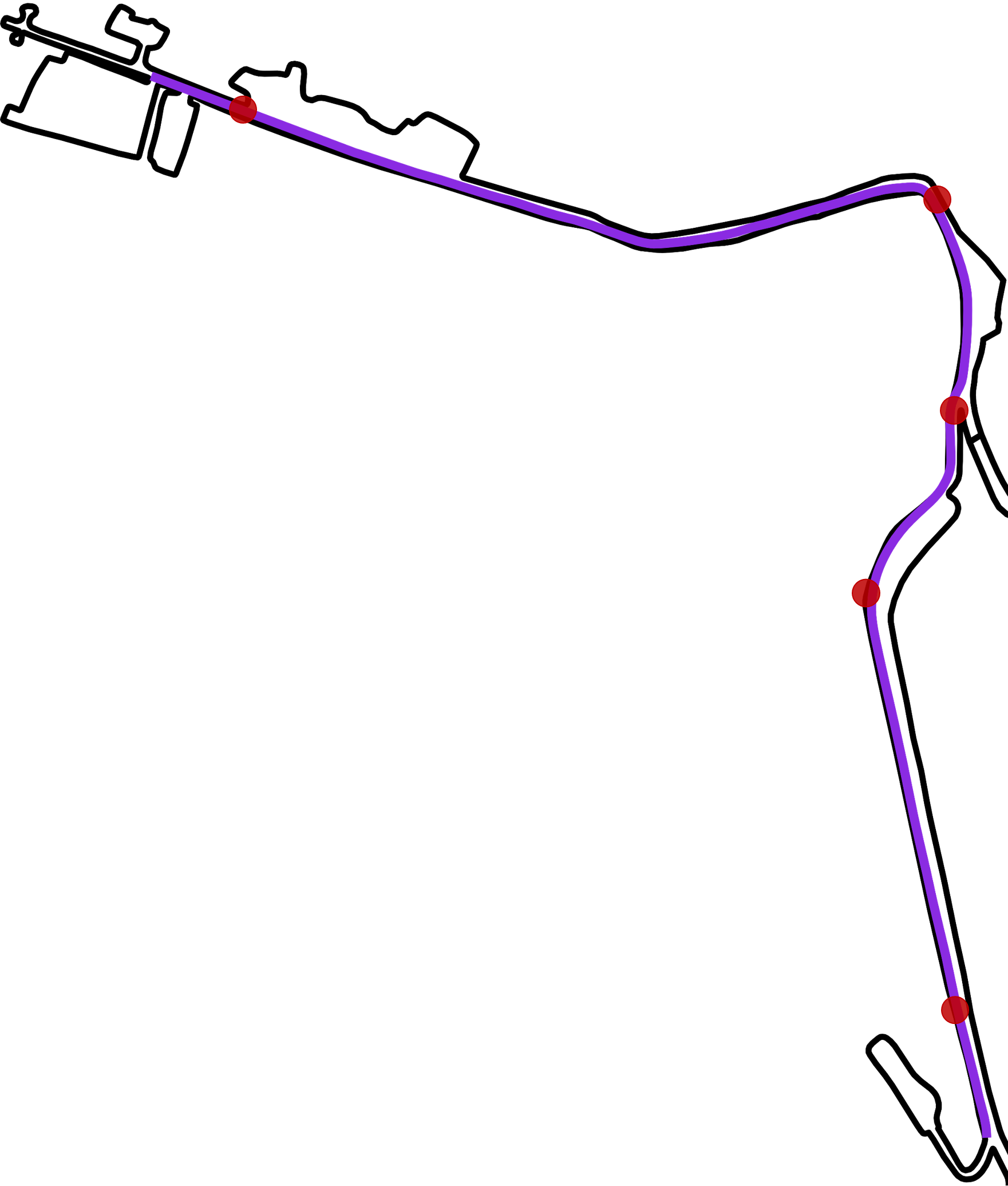}}

\subfigure[Instantaneous with counter-clockwise]{
\label{Fig:sub.2}
\includegraphics[width=0.2\textwidth]{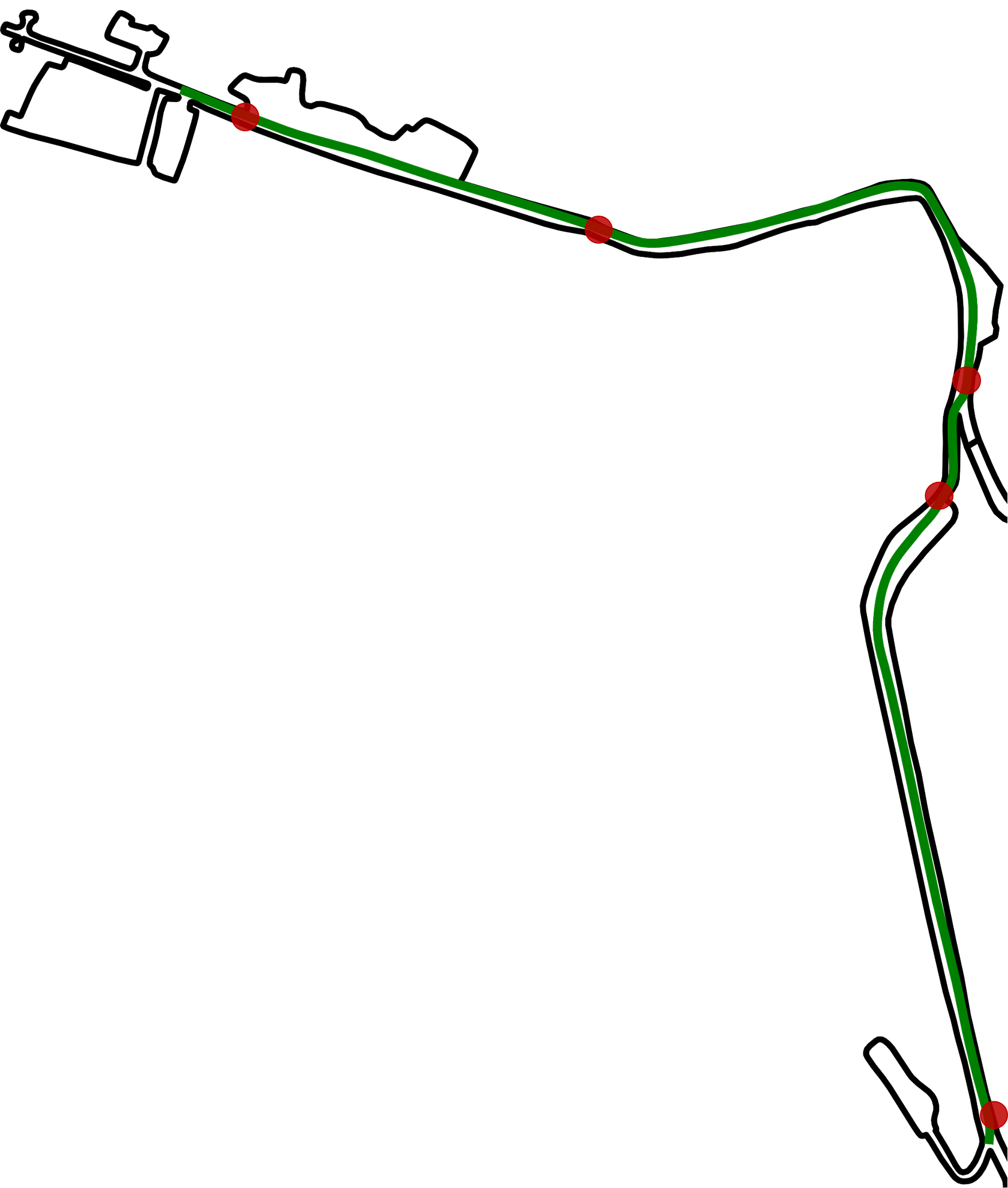}}\hspace{3mm}
\subfigure[Uniform fusion with counter-clockwise]{
\label{Fig:sub.2}
\includegraphics[width=0.2\textwidth]{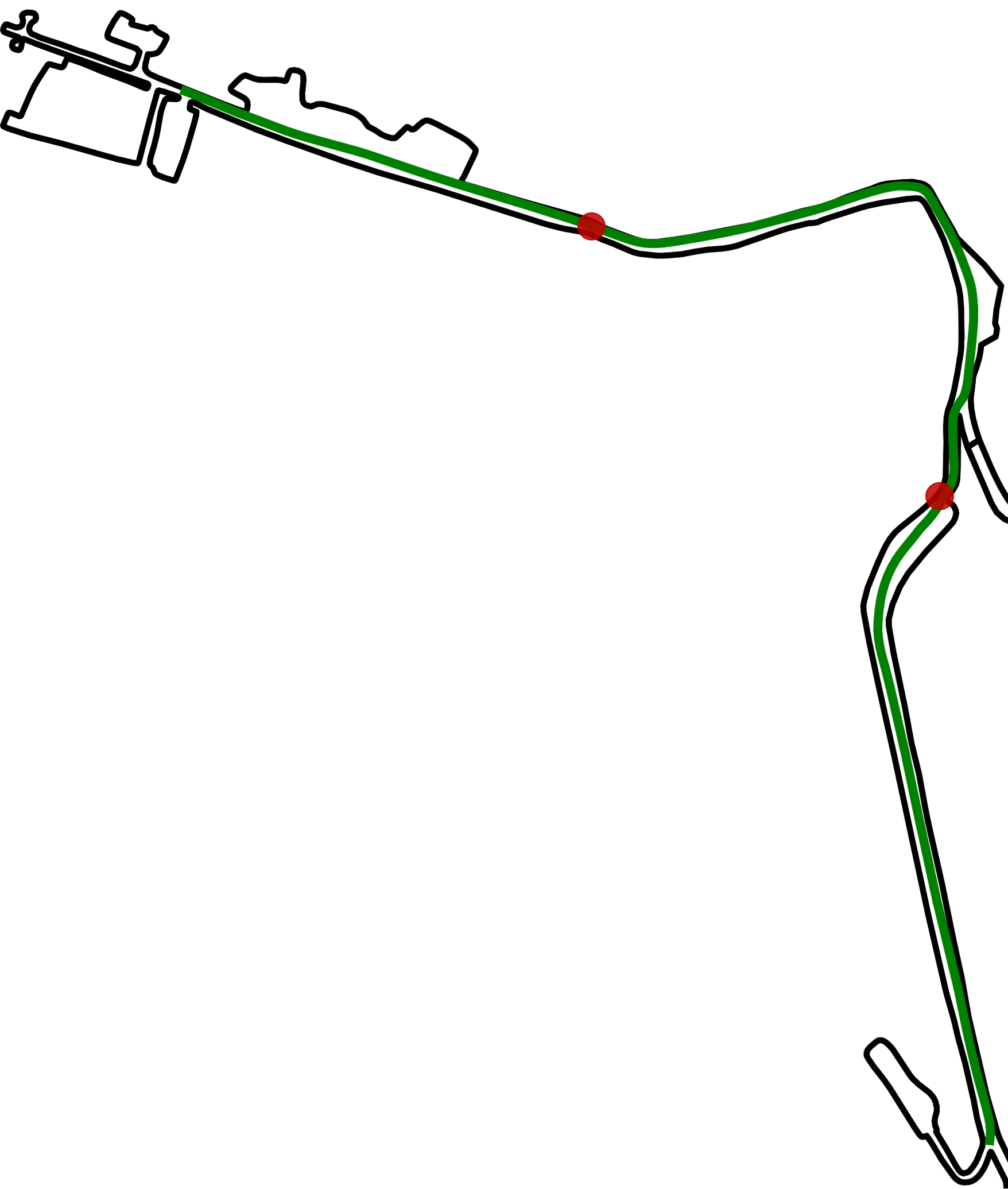}}\hspace{3mm}
\subfigure[Evidential fusion with counter-clockwise]{
\label{Fig: E1F}
\includegraphics[width=0.2\textwidth]{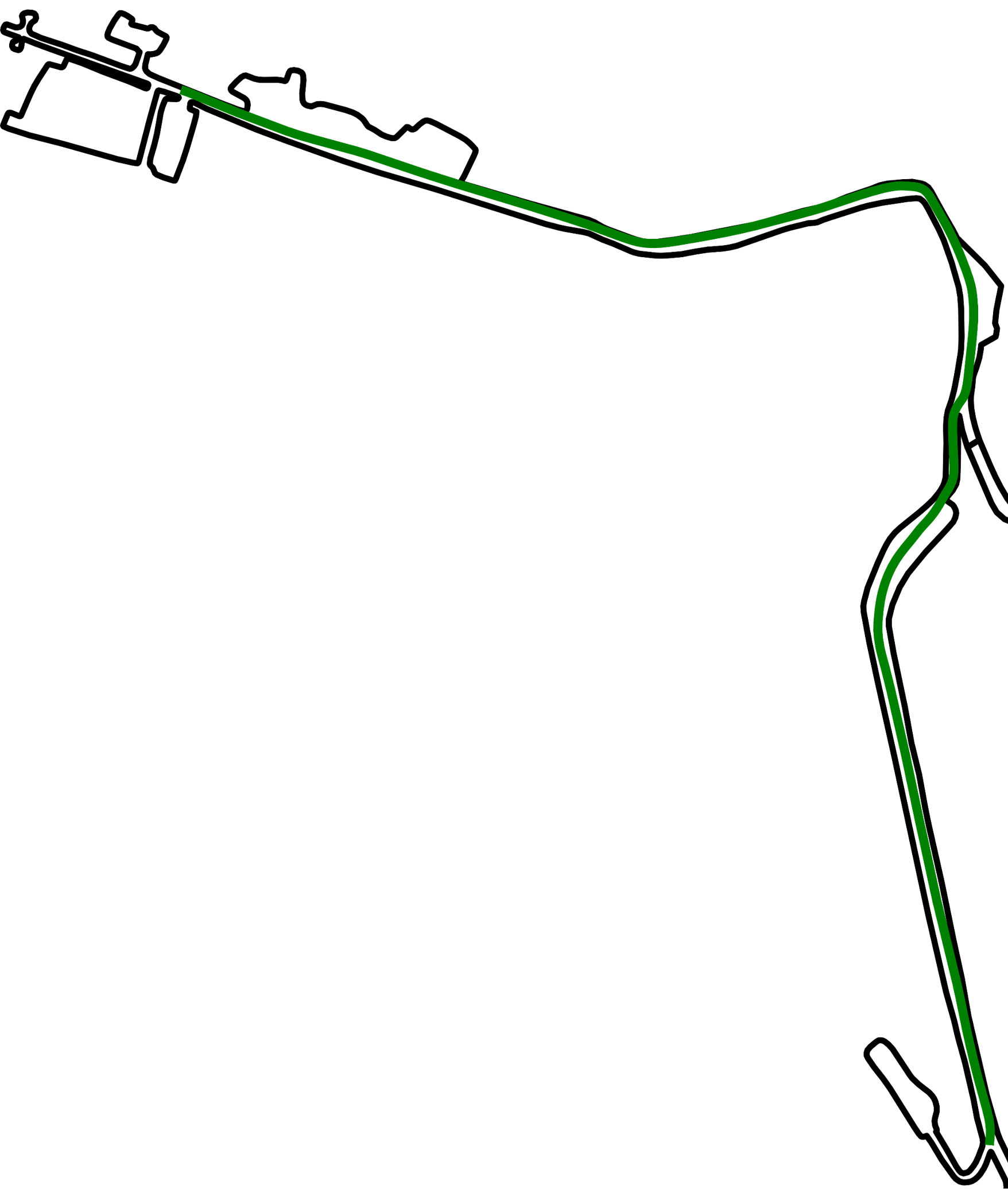}}
\subfigure[Evidential fusion with counter-clockwise in OOD]{
\label{Fig:gpsOODcounterclockwise}
\includegraphics[width=0.2\textwidth]{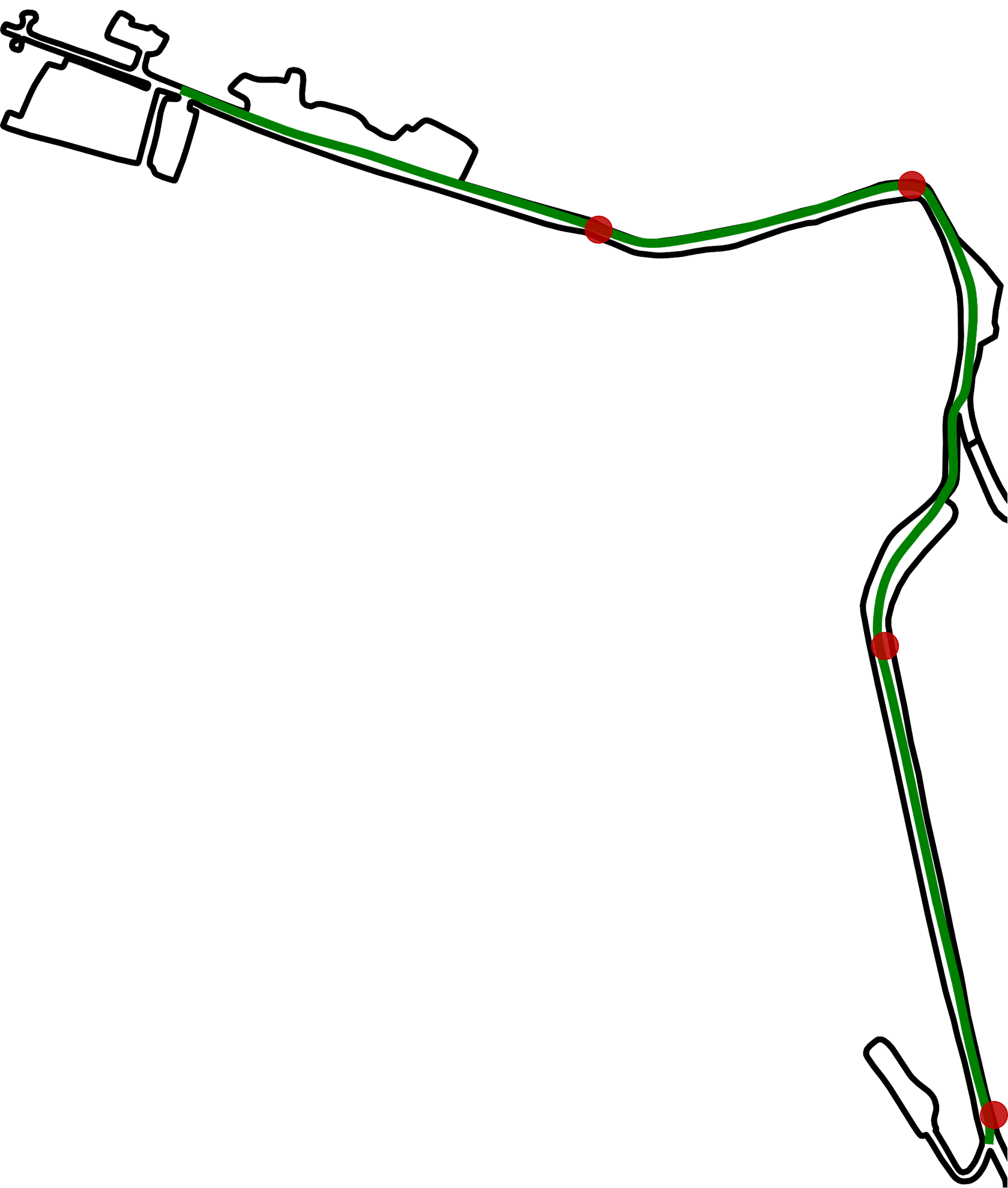}}

\caption{The verification results of the three modes of FusionPlanner without sensor failures. In the majority of open-pit mines, transportation roads are bi-directional, with varying constraints on sensor perception during left and right turns. In this experiment, a separate comparison is presented. The first row corresponds to clockwise driving, while the second is counter-clockwise driving. The red dots mark collision or intervention. The red dots represent collision or intervention.}
\label{Fig.lane-keep}
\end{figure}

We first validate the performance of our well-trained motion planner in the lane-keep task on a 1500 m two-way transportation road in the open-pit mine. As shown in Fig.\ref{Fig.lane-keep}, the counterclockwise (green) trajectories exhibit higher robustness compared to the clockwise (purple) trajectories. This difference may stem from that during left turns, the LiDAR‘s perception is less obstructed by retaining walls compared to right turns, thereby more comprehensive perception receives. Consequently, FusionPlanner demonstrates superior overall robustness during left turns in comparison to right turns. The experiment also compared the three control modes of FusionPlanner under normal sensor conditions. Evidential fusion model achieves the most robust results, even enabling counterclockwise driving without intervention (Fig.\ref{Fig: E1F}). Furthermore, the robustness of uniform fusion model is improved compared to instantaneous model, demonstrating that leveraging historical commands indeed enhances the control stability of the truck.

Open-pit mines are always located in remote areas, and occurrences of GNSS signal interruptions are not uncommon. In this task, to evaluate the trustworthiness and robustness of FusionPlanner in OOD events, we intentionally triggered sensor failures in the localization layer with a 4$\%$ probability. In Fig.\ref{Fig:gpsOODclockwise} and Fig.\ref{Fig:gpsOODcounterclockwise}, we tested evidential fusion model in this adverse situation.  Compared with the normal state (Fig.\ref{Fig:clocknormal} and Fig.\ref{Fig: E1F}), the number of red dots increased in both driving directions. The trucks are stuck into OOD events due to signal loss in a particular frame, which subsequently leads to irrecoverable issues and ultimately results in collisions or intervention.




\subsection{Disturbance tasks}
The recovery from an abnormal state demonstrates the robustness of our planner when encountering OOD events. In this evaluation task, we artificially distorted the initial state of the truck to simulate abnormal conditions. A stochastic error $\delta_{yaw} \in [-10^{\circ },+10^{\circ }]$ is added to the initial heading angle, along with a random lateral Offset $\delta_{coor_x} \in [-1m, +1m]$ applied to the initial localization. Successful recovery is marked by the truck returning to the reference line within 20s, while collision or failure to return to a safe state within the specified time is considered as unsuccessful. As presented in Tab. \ref{tab: covering}, both uniform fusion and evidential fusion show better self-correcting capabilities. All models achieve higher records from turning left than turning right which might be because LiDAR achieves an improved field of vision, minimizing the occlusion of crucial information.

\linespread{1.05}
\begin{table}[H]  \centering \footnotesize
\caption{Performance of recovering from the distorted state. Here, Straight denotes the success rate of recovering to the reference line (safe state) in a straight road, and Turn Left and Turn Right demonstrate in intersection roads, respectively.}
\label{tab: covering}
\begin{tabular}{p{2.5cm}<{\centering}|p{2.5cm}<{\centering} p{2.5cm}<{\centering} p{2.5cm}<{\centering} p{2.5cm}<{\centering}}
\hline
\rowcolor[HTML]{EFEFEF} 
Model    & Stright   & Turn Left & Turn Right & Average     \\ \hline
Instance   & 0.75 & 0.63 & 0.44  & 0.667   \\
Uniform Fusion  & 0.89 & 0.73 & 0.59  & 0.737  \\
Evidence Fusion & 0.90 & 0.71 & 0.61  & 0.74 \\ \hline
\end{tabular}
\end{table}
\subsubsection{Navigation Task}
\begin{figure}[h]
\centering 
\subfigure[The transportation trajectory in clockwise navigation tasks.]{
\label{Fig:navtasksub.1}
\includegraphics[width=0.45\textwidth]{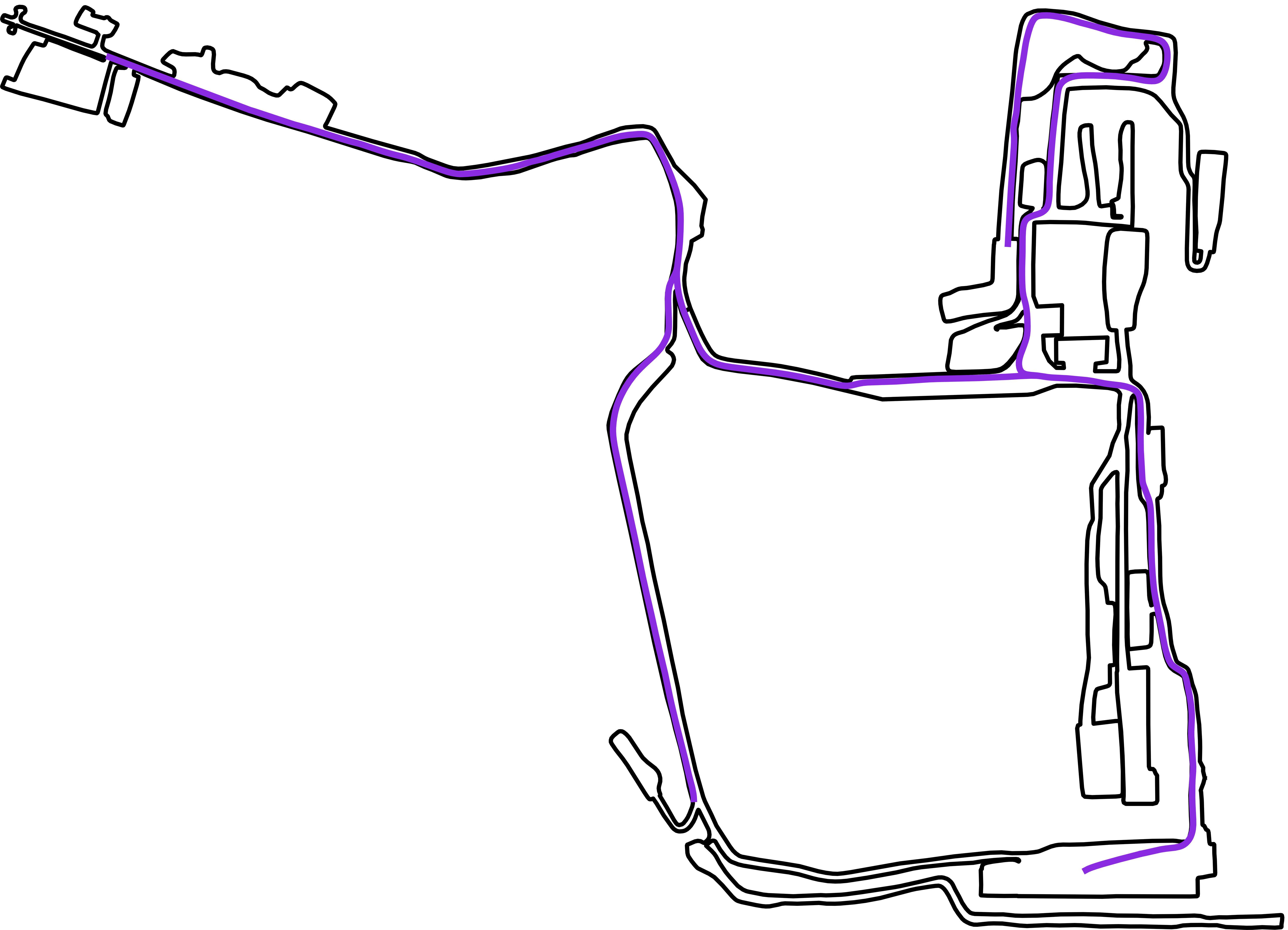}}
\subfigure[The transportation trajectory in counter-clockwise navigation tasks.]{
\label{Fig:navtasksub.2}
\includegraphics[width=0.45\textwidth]{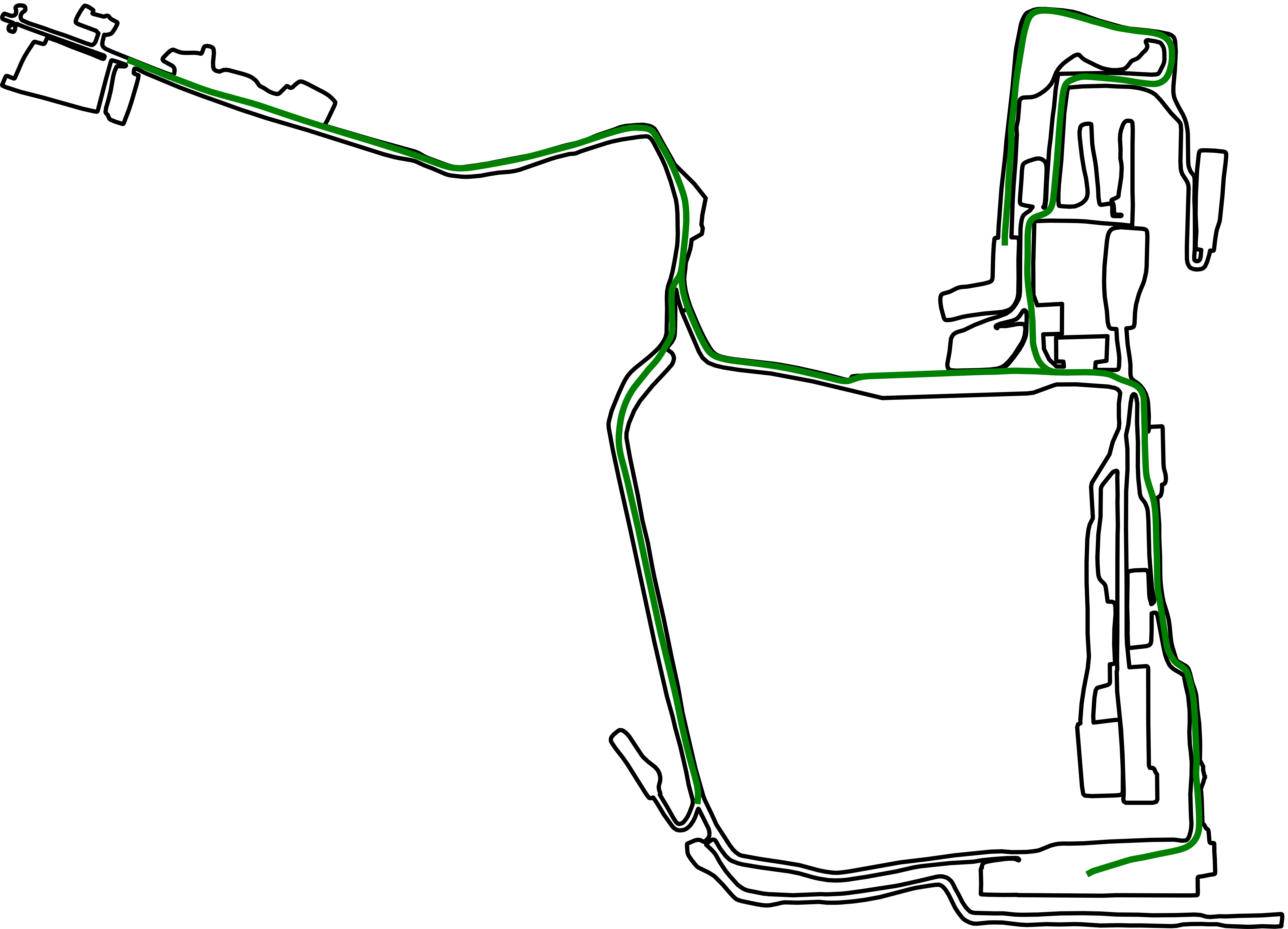}}

\caption{The transportation trajectory with 11,000m in navigation tasks.}
\label{Fig.navtask}
\end{figure}

\begin{figure}[t]
\centering  
\subfigure[]{
\label{Fig:navtasksub.1}
\includegraphics[width=0.45\textwidth]{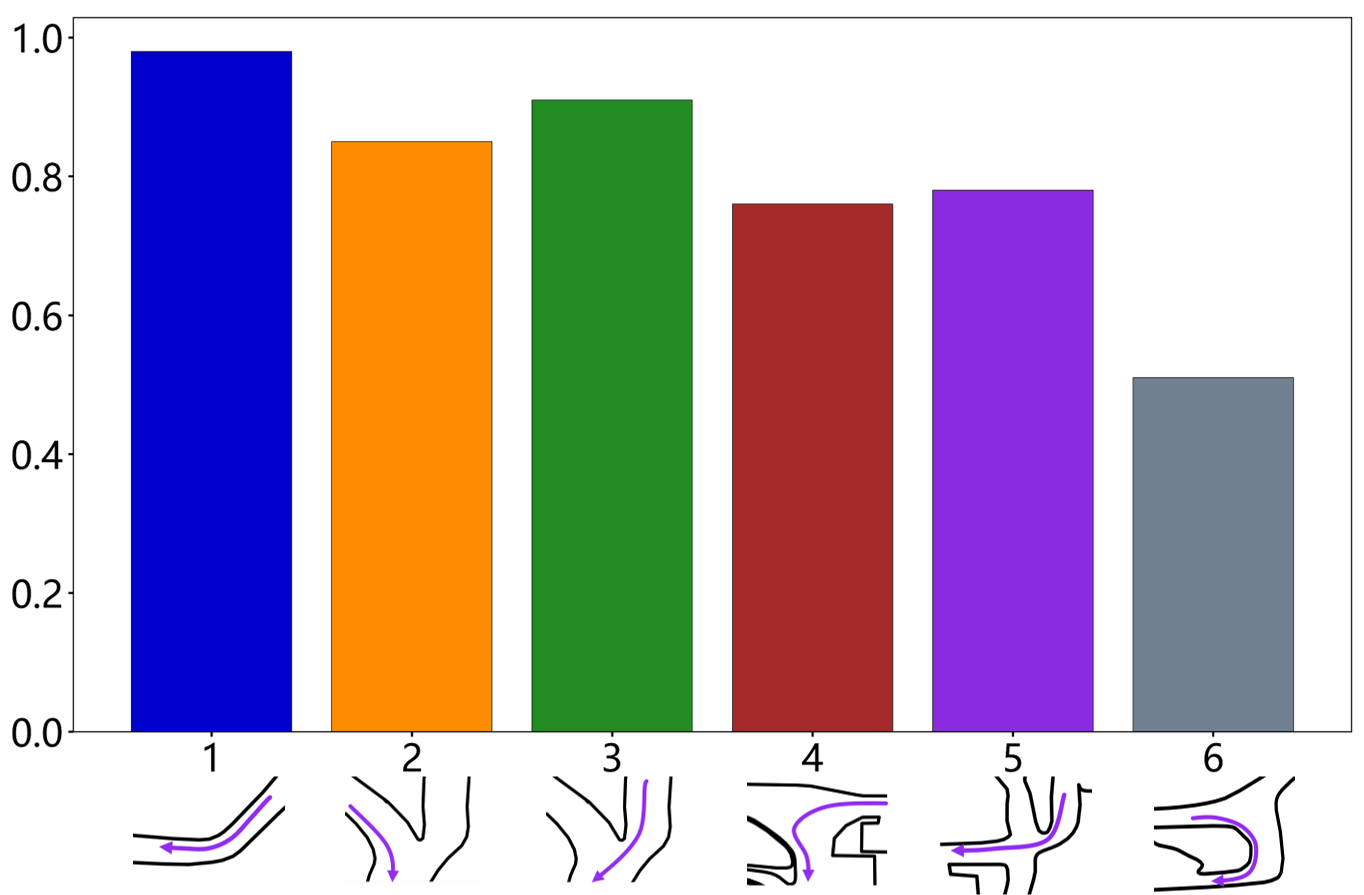}}
\subfigure[]{
\label{Fig:navtasksub.2}
\includegraphics[width=0.45\textwidth]{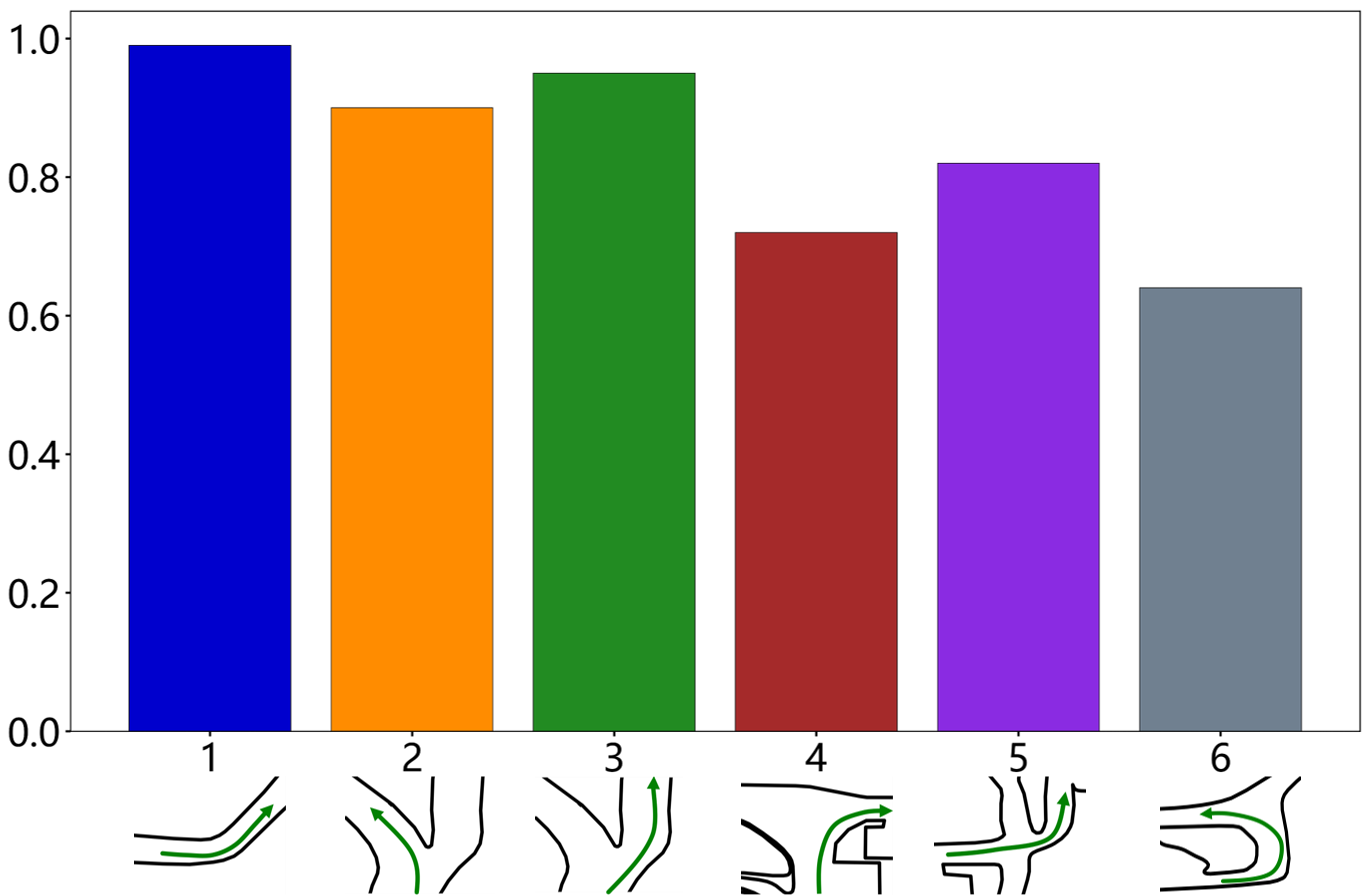}}

\caption{Success rates at individual intersections in the open-pit mine.}
\label{Fig.navtaskintersection}
\end{figure}

The navigation task serves as a comprehensive validation task for autonomous driving algorithms. In line with the MiningNav benchmark, we conduct evaluations of FusionPlanner on two-way transportation roads spanning a length of 11000 m. Each testing iteration consists of a minimum of 1000 m of path and includes at least one turning intersection, the clockwise and counter-clockwise transportation is demonstrated in Fig.\ref{Fig.navtask}. Our primary focus was to evaluate the performance of FusionPlanner at different intersections along the transportation roads, considering both clockwise and counterclockwise directions. As shown in Fig.\ref{Fig.navtaskintersection}, the truck demonstrates a higher success rate when driving into or out of smoother intersections (such as intersections 1, 2, 3, and 5), while encountering lower success rates at sharper intersections (such as intersections 4 and 6). It is important to note that intersection 6, with its long trajectory and sharper angle, posed a particular challenge and resulted in less satisfactory outcomes for our planner. Furthermore, similar to the lane-keeping tasks, the success rate of left turns (indicated in green) was generally higher than that of right turns (indicated in purple).

\section{Conclusion}

Autonomous driving has been studied since the mid-20th century. After almost 70 years of research and investigation, significant achievements have been accomplished. The traditional pipeline framework is found in widespread implementation in the industry, and the novel end-to-end planning framework arising significant attention in the academic community. One of its primary advantages lies in the ability of deep models to emulate the nonlinear decision-making of human drivers. However, most of the research is concentrated on structured urban environments, neglecting unstructured scenarios such as open-pit mines, which pose distinctive challenges for autonomous driving and are a longstanding issue for both academia and industry.

\textcolor{black}{In this research, we propose a comprehensive paradigm for unmanned transportation in open-pit mines, taking into account the complex operational conditions and harsh environmental factors. Firstly, we introduce an end-to-end motion planning model, called FusionPlanner, specifically designed for mining trucks. FusionPlanner incorporates various methods such as optimizing sparse convolution kernels for fast point cloud processing, employing multi-task decoupling of lateral and longitudinal control commands for stable control, and integrating an evidential fusion model at the execution level to fuse predictive control commands from multiple frames. Next, the first benchmark for unmanned transportation, named MiningNav is proposed, which includes three customized tasks designed to comprehensively assess the robustness of the algorithms in challenging mining scenarios. Finally, we construct a simulation model, named Parallel Mining Simulator (PMS), specifically designed for open-pit mines. PMS constructs a comprehensive, high-fidelity virtual environment, serving as a simulation platform for the training and testing of related algorithms. Through experiments in PMS, FusionPlanner demonstrates stable navigation capabilities, effectively reducing the probability of collisions or interventions in open-pit mining scenarios. FusionPlanner takes initial steps toward achieving end-to-end autonomous driving in open-pit mines. Yet, in some OOD events, planning failures still occur. Improving the trustworthiness and robustness of the planner is a field that needs further attention. Moreover, the involvement of foundation models \cite{bigmodel} introduces new possibilities for end-to-end motion planning.}

\bibliographystyle{elsarticle-num}
\bibliography{MyRef}







\end{document}